\pdfoutput=1
\documentclass[10pt,twocolumn,letterpaper]{article}
\usepackage[pagenumbers]{cvpr} %
\usepackage{cuted}
\usepackage{currfile}
\usepackage{caption}
\usepackage{graphicx}
\usepackage{amsmath}
\usepackage{amssymb}
\usepackage{booktabs}
\usepackage{placeins}
\usepackage{enumitem}
\usepackage{multirow}
\usepackage{xspace}
\xspaceaddexceptions{\% \, $}
\usepackage[table]{xcolor}
\usepackage{colortbl}
\definecolor{oursbg}{RGB}{226,238,252}

\newcommand{\method}{OccJEPA\xspace}
\newcommand*{\projecturl}{}
\newcommand{\nSvOccOurs}{43.68\xspace}
\newcommand{\nSvMiouOurs}{78.60\xspace}
\newcommand{\nSvOffOccOurs}{52.09\xspace}
\newcommand{\nSvKeepOurs}{51.4\%\xspace}
\newcommand{\nSvOccDvas}{39.98\xspace}
\newcommand{\nSvMiouDvas}{77.08\xspace}
\newcommand{\nSvOffOccDvas}{55.12\xspace}
\newcommand{\nSvKeepDvas}{44.5\%\xspace}
\newcommand{\nSvOccConvex}{20.75\xspace}

\newcommand{\nSvOccDiff}{+3.70\xspace}
\newcommand{\nSvOccDiffCI}{[+1.38, +5.40]\xspace}
\newcommand{\nSvMiouDiff}{+1.52\xspace}
\newcommand{\nSvMiouDiffCI}{[+0.73, +2.33]\xspace}
\newcommand{\nSvOffOccDiff}{$-$3.03\xspace}
\newcommand{\nSvOffOccDiffCI}{[$-$5.26, $-$0.74]\xspace}

\newcommand{\nSvFamN}{23\xspace}
\newcommand{\nDvasMissedPct}{27.5\%\xspace}
\newcommand{\nSvNseq}{5,432\xspace}
\newcommand{\nSvNvid}{39\xspace}

\newcommand{\nSvNset}{83,245\xspace}

\newcommand{\nTaoAPDvas}{89.25\xspace}

\newcommand{\nTaoAPDiff}{+4.11\xspace}

\newcommand{\nTaoNframes}{34,800\xspace}
\newcommand{\nTaoNvid}{542\xspace}

\newcommand{\nTtAPDiff}{+4.00\xspace}

\newcommand{\nTtLowVisDiff}{+29.44\xspace}
\newcommand{\nTtLowVisDiffCI}{[+21.47, +36.70]\xspace}
\newcommand{\nTtNframes}{48,675\xspace}
\newcommand{\nTtNvid}{772\xspace}
\newcommand{\nSamAPOurs}{85.59\xspace}
\newcommand{\nSamAPDvas}{81.27\xspace}
\newcommand{\nSamAPDiff}{+4.32\xspace}
\newcommand{\nSamAPDiffCI}{[+3.37, +5.34]\xspace}
\newcommand{\nSamLowVisDiff}{+22.22\xspace}
\newcommand{\nSamIoU}{0.83\xspace}

\newcommand{\nMovidOurs}{84.66\xspace}
\newcommand{\nMovidDvas}{82.97\xspace}
\newcommand{\nMovibOurs}{88.94\xspace}
\newcommand{\nMovibDvas}{87.80\xspace}
\newcommand{\nMovidOccDiff}{+4.14\xspace}
\newcommand{\nMovibOccDiff}{+3.31\xspace}

\newcommand{\nSvRepMiouDvas}{77.07\xspace}
\newcommand{\nSvRepOccDvas}{55.12\xspace}
\newcommand{\nKubSDiff}{+1.19\xspace}

\newcommand{\nKubDDiff}{+3.57\xspace}

\newcommand{\nKubDOccDiff}{+7.71\xspace}
\newcommand{\nSpOurs}{0.352\xspace}
\newcommand{\nSpDvas}{11.77\xspace}
\newcommand{\nSpXClip}{33\xspace}
\newcommand{\nSpOursMs}{14.6\xspace}
\newcommand{\nSpDvasMs}{471\xspace}
\newcommand{\nSpX}{32\xspace}
\newcommand{\nCostOursFrameTF}{1.48\xspace}
\newcommand{\nCostDvasFrameTF}{36.1\xspace}
\newcommand{\nCostRatio}{24\xspace}
\newcommand{\nWmLFrameTF}{0.152\xspace}
\newcommand{\nWmLRatio}{238\xspace}

\newcommand{\nSpOursSeq}{0.70\xspace}
\newcommand{\nSpXSeq}{17\xspace}
\newcommand{\nCostOursSeqTF}{71.1\xspace}
\newcommand{\nCostRatioSeq}{13\xspace}
\newcommand{\nSpDepth}{3.2\xspace}
\newcommand{\nSpDiff}{8.6\xspace}
\newcommand{\nSvPredAll}{+8.53\xspace}

\newcommand{\nTaoPredAll}{+3.93\xspace}

\newcommand{\nSvPredCap}{+2.15\xspace}

\newcommand{\nTaoPredCap}{+0.74\xspace}

\newcommand{\nShareSv}{75\%\xspace}
\newcommand{\nShareTao}{81\%\xspace}
\newcommand{\nTtPred}{+4.08\xspace}

\newcommand{\nMGain}{+9.83\xspace}

\newcommand{\nMW}{+7.41\xspace}

\newcommand{\nGW}{+5.72\xspace}

\newcommand{\nMVsL}{$-$2.57\xspace}

\newcommand{\nMVsLTao}{$-$0.84\xspace}

\newcommand{\nGVsG}{+1.43\xspace}

\newcommand{\nMEncSv}{26.75\xspace}
\newcommand{\nLEncSv}{30.73\xspace}
\newcommand{\nGainB}{+8.67\xspace}

\newcommand{\nGainG}{+8.19\xspace}
\newcommand{\nRcSv}{+0.36\xspace}

\newcommand{\nRcTao}{+1.31\xspace}

\newcommand{\nRcLowVis}{+11.33\xspace}

\newcommand{\nCostOursTF}{35.6\xspace}
\newcommand{\nCostDvasTF}{903\xspace}
\newcommand{\nCostRatioClip}{25\xspace}
\newcommand{\nParamsOurs}{1.91B\xspace}
\newcommand{\nParamsDvas}{2.59B\xspace}
\newcommand{\nTrainOurs}{69M\xspace}
\newcommand{\nParamsDvasUnet}{1.52B\xspace}
\newcommand{\nWmLTF}{3.64\xspace}
\newcommand{\nWmLGap}{0.83\xspace}

\newcommand{\nWmBTao}{90.13\xspace}
\newcommand{\nWmLTaoDiff}{+1.83\xspace}
\newcommand{\nWmLTaoDiffCI}{[+0.83, +2.91]\xspace}
\newcommand{\nTrainHours}{8\xspace}
\newcommand{\nTrainMem}{17.0\xspace}
\newcommand{\nNoneSvOcc}{35.15\xspace}
\newcommand{\nNoneTao}{89.43\xspace}
\newcommand{\nOcSvLowOurs}{30.9\xspace}
\newcommand{\nOcSvLowDvas}{43.3\xspace}
\newcommand{\nOcSvHighOurs}{39.7\xspace}
\newcommand{\nOcSvHighDvas}{30.6\xspace}
\newcommand{\nOcSvGapMin}{6.3\xspace}
\newcommand{\nOcSvGapMax}{9.1\xspace}

\newcommand{\nOcTaoQDvas}{24.5\xspace}

\newcommand{\nOcTaoHDvas}{43.4\xspace}

\newcommand{\nIbSeqs}{270\xspace}
\newcommand{\nIbOurF}{66.5\%\xspace}
\newcommand{\nIbOurZ}{45.8\%\xspace}
\newcommand{\nIbRndF}{6.6\%\xspace}
\newcommand{\nIbRndZ}{61.8\%\xspace}
\newcommand{\nIbRelF}{11.4\%\xspace}

\newcommand{\nIbOurBg}{3.6\%\xspace}

\newcommand{\nIaGainB}{+18.16\xspace}
\newcommand{\nIaGainC}{+13.34\xspace}
\newcommand{\nTsDevOurs}{7.15\xspace}

\newcommand{\nTsDevDvas}{3.96\xspace}

\renewcommand{\dbltopfraction}{0.85}
\renewcommand{\topfraction}{0.85}
\renewcommand{\textfraction}{0.1}
\renewcommand{\floatpagefraction}{0.8}
\renewcommand{\dblfloatpagefraction}{0.8}

\makeatletter\let\supp@maketitle\@maketitle\makeatother   %
\makeatletter   %
\patchcmd{\@maketitle}{\vskip .375in}{\vskip .178in}{}{\PackageWarning{main}{title patch failed}}
\makeatother
\definecolor{cvprblue}{rgb}{0.21,0.49,0.74}
\usepackage[pagebackref,breaklinks,colorlinks,allcolors=cvprblue]{hyperref}
\usepackage[capitalize]{cleveref}
\usepackage{dblfloatfix}      %

\def\confName{CVPR}
\def\confYear{2027}

\title{OccJEPA: Occlusion as Masking for Amodal Video Segmentation}
\author{%
Genpei Zhang\textsuperscript{1}\quad
Runmin Jiang\textsuperscript{2}\quad
Xingzu Zhan\textsuperscript{3}\quad
Yibo Zhong\textsuperscript{1}\quad
Ruien Li\textsuperscript{1}\quad
Zhou Zhang\textsuperscript{1}\\
\textsuperscript{1}University of Wisconsin--Madison\quad
\textsuperscript{2}Carnegie Mellon University\quad
\textsuperscript{3}Johns Hopkins University%
}

\begin{document}
\maketitle
\begin{abstract}
Amodal video segmentation asks for the complete mask of an object in every frame, including the parts hidden behind
other objects. The best current methods fine-tune video diffusion models to generate these masks; they need a
depth network and tens of denoising steps, and their outputs vary from sample to sample. Self-supervised video world
models of the JEPA family are pre-trained on a closely related task, predicting the representations of masked
space-time regions from the visible ones, yet their predictor is usually discarded after pre-training. We reuse it by treating occlusion as masking. Tokens not covered by the object's
visible mask are dropped, the V-JEPA~2.1 predictor is fine-tuned to fill them in, and a lightweight
decoder turns all features into amodal masks. The resulting model, \textbf{OccJEPA} (\textbf{Occ}lusion-as-masking
\textbf{JEPA}), is deterministic and \nSpX$\times$ faster per frame than Diffusion-VAS, the diffusion-based state of the art. It
is also more accurate on SAIL-VOS (\nSvOccDiff occluded-region mIoU over all occluded frames), on TAO-Amodal
(\nTtAPDiff AP$_{50}$ on a held-out test split, \nSamAPDiff with SAM2-tracked masks) and on Kubric. Controlled
experiments trace most of the gain to the predictor's pre-trained weights rather than its capacity, and the predictor
helps at every encoder size tested (87M to 1.85B parameters).
\ifx\projecturl\empty\else Code is available at \url{\projecturl}.\fi
\end{abstract}

\begin{figure}[t]
\centering
\includegraphics{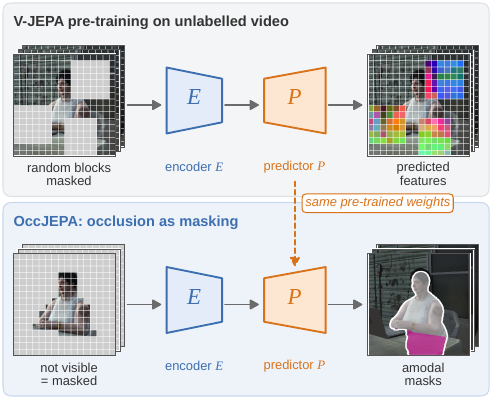}
\caption{\textbf{Concept.} In pre-training without labels (top), the V-JEPA encoder $E$ reads the visible tokens of a clip, and the predictor $P$ infers the features of the masked blocks (colours show the first three principal components).
\method (bottom) masks the tokens not covered by the object's visible mask, and the same predictor, fine-tuned, fills them in. A lightweight decoder turns the result into the amodal mask, with the completed part in magenta. Real SAIL-VOS data;
the top block mask is illustrative.}
\label{fig:concept}
\end{figure}
\section{Introduction}
\label{sec:intro}

When a parked car hides a pedestrian's legs, people still see a whole person, and trackers, robots and video editors need the same ability. Amodal segmentation~\cite{li2016amodal,zhu2017semantic} formalises it as
predicting the complete mask of an object, visible and hidden parts together. In video~\cite{sailvos,taoamodal}, the
object's visible mask is given in every frame, and its full extent must be recovered over time.

Completing a hidden region requires a prior on how objects are shaped and how they continue behind occluders. Earlier
methods learn it from amodal annotations~\cite{aisformer,bcnet} or synthetic occlusions~\cite{pcnet,c2fseg}, and video
methods add temporal cues~\cite{savos,eoras}. The best current results come from generative models. Diffusion-VAS~\cite{dvas}
fine-tunes Stable Video Diffusion~\cite{svd} to generate amodal mask videos from the visible masks and estimated
depth~\cite{depthanythingv2}. This generative prior is strong, but each clip costs a depth network and tens of denoising steps,
and different samples give different answers.

Self-supervised video world models offer another prior. Joint-embedding predictive architectures
(JEPAs)~\cite{lecun2022path,ijepa} hide large space-time regions of a clip and train a \emph{predictor} to infer their
representations from the visible remainder~\cite{vjepa,vjepa2,vjepa21}. Hidden regions often contain parts of
objects, so the predictor must learn how objects continue. Amodal completion asks a closely related question about the hidden part of a single object. Yet the predictor is usually discarded after
pre-training.

\begin{figure*}[t]
  \centering
  \includegraphics[width=\textwidth]{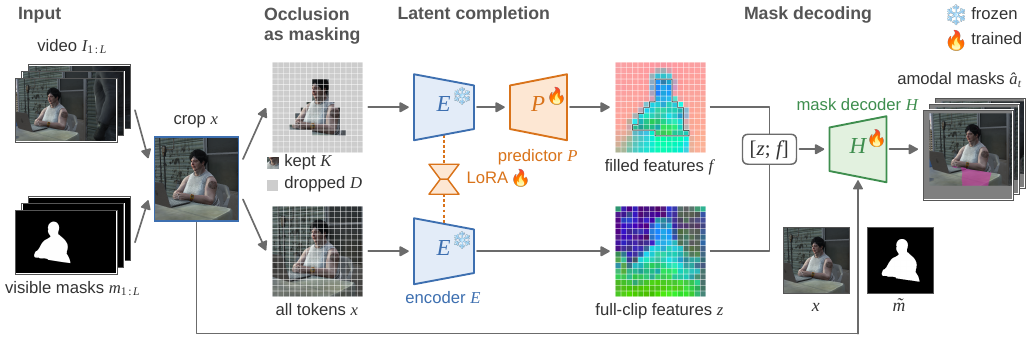}
  \caption{\textbf{Overview of \method.} The clip $x$ is cropped around the visible mask (\cref{eq:box}), and tokens
  with visible coverage below $\tau$ are dropped (\cref{eq:cover}). The V-JEPA~2.1 encoder $E$ reads the kept tokens,
  and its pre-trained predictor $P$, fine-tuned, fills in the dropped ones ($f$, \cref{eq:fill}); $E$ also reads the full clip ($z$)
  and is frozen except for one shared LoRA adapter. The mask decoder $H$ maps $[z;f]$, $x$ and $\tilde m$ to amodal
  masks (\cref{eq:paste}). Real SAIL-VOS data. Feature maps show the first three principal components, black outlines mark the kept tokens, and magenta marks the completed part.}
  \label{fig:method}
\end{figure*}

We therefore treat \emph{occlusion as masking} (\cref{fig:concept,fig:method}). Tokens not covered by the object's
visible mask are dropped, as masked tokens are in pre-training. The pre-trained V-JEPA~2.1 predictor, fine-tuned for
the task, predicts their features from the visible tokens alone, and a lightweight decoder maps all features to amodal
masks. The dropped tokens also cover occluders and background, so no occluder mask is needed, and the model itself decides where the object continues. We call the model \textbf{OccJEPA} (\textbf{Occ}lusion-as-masking \textbf{JEPA}); it
needs neither depth nor sampling. Whether the predictor's prior transfers is not obvious, because pre-training asks it what a masked region shows, occluders included, and not what lies behind them.

On SAIL-VOS~\cite{sailvos}, \method reaches an occluded-region mIoU of \nSvOccOurs over all occluded frames, against
\nSvOccDvas for Diffusion-VAS. On the real videos of TAO-Amodal~\cite{taoamodal}, it improves AP$_{50}$ on val
(\nTaoAPDiff) and on a held-out test split built the same way (\nTtAPDiff), and it keeps its lead
when SAM2~\cite{sam2} tracks the visible masks. It also outperforms Diffusion-VAS on the Kubric benchmarks~\cite{taco}
and runs \nSpX$\times$ faster per frame. Controlled experiments show that the prior does transfer. The predictor's pre-trained weights account for \nShareSv (SAIL-VOS) and \nShareTao (TAO-Amodal) of what the
predictor adds and move the completion into the predictor. It helps at every encoder size tested
(87M to 1.85B parameters).

We make three contributions.
\begin{itemize}[leftmargin=*,itemsep=1pt,topsep=2pt,parsep=0pt]
  \item We cast amodal video segmentation as masked latent prediction, \emph{occlusion as masking}, with the released predictor of a self-supervised video world model.
  \item We build \method, a fast, deterministic model that, with one set of weights, outperforms Diffusion-VAS on
  SAIL-VOS, TAO-Amodal and Kubric.
  \item We show with controlled experiments and a feature-space analysis that the gain comes mainly from the
  predictor's pre-trained weights, not its capacity. Fine-tuned from them, the predictor fills \nIbOurF of the
  hidden-object tokens with object features, against \nIbRndF from scratch.
\end{itemize}

\section{Related Work}
\label{sec:related}

\paragraph{Amodal segmentation.}
Li and Malik~\cite{li2016amodal} introduced amodal instance segmentation, and COCOA~\cite{zhu2017semantic} and
KINS~\cite{kins} provide human annotations for it. Image models use occlusion-aware heads~\cite{bcnet,aisformer}, shape
priors~\cite{vrsp,c2fseg}, synthetic de-occlusion~\cite{pcnet} or layered scene completion~\cite{zheng2021visiting}, as surveyed in~\cite{amodalsurvey}. Recent methods borrow priors from image
diffusion~\cite{pix2gestalt,pmcd,sdamodal}, inpainting~\cite{tuningfree} and segmentation foundation
models~\cite{sameo,amodalsam,sam,sam2}. In video, SAIL-VOS~\cite{sailvos,sailvos3d} and TAO-Amodal~\cite{taoamodal} provide
amodal annotations for synthetic and real clips. SaVos~\cite{savos} learns from motion without amodal labels,
EoRaS~\cite{eoras} uses object-centric bird's-eye-view representations, A2VIS~\cite{a2vis} adds amodal awareness to video
instance segmentation, and several trackers keep objects through occlusion~\cite{permatrack,randomwalkperm,tcow,tabe}.
Diffusion-VAS~\cite{dvas} fine-tunes Stable Video Diffusion~\cite{svd} to generate amodal masks from visible masks and
pseudo-depth~\cite{depthanythingv2}, TACO~\cite{taco} completes the hidden RGB content with a video diffusion model, and
IC-Amodal~\cite{icamodal} guides video amodal completion with in-context exemplars. \method instead completes the object
in the latent space of a predictive model, deterministically and without depth.

\paragraph{Self-supervised video models and world models.}
Masked modelling reconstructs the pixels~\cite{mae,videomae,videomaev2} or the tokens and features~\cite{beit,maskfeat,data2vec}
of masked regions; generative imputation likewise completes partially observed time series~\cite{misspath}. JEPAs
instead predict the representations of masked regions in a learned latent
space~\cite{lecun2022path,ijepa}. V-JEPA~\cite{vjepa} does so for video with large space-time masks, V-JEPA~2~\cite{vjepa2}
scales it and adds an action-conditioned predictor for planning, and V-JEPA~2.1~\cite{vjepa21} adds dense objectives.
Planning with latent world models~\cite{worldmodels,dinowm} uses the predictor, but downstream perception usually
keeps only the encoder. V-JEPA's prediction error reveals an understanding of object permanence~\cite{intuitivephysics}, and
concurrent work asks what video world models retain about a hidden object~\cite{notpermanence}. Closest to ours, a JEPA
predictor completes occluded features for image classification~\cite{jepapredictor}; we use a video predictor as the
completion module of an amodal video segmenter, evaluated on standard benchmarks.

\paragraph{Repurposing pre-trained models.}
Generative models have been adapted to depth~\cite{marigold,depthcrafter} and semantic tasks~\cite{vpd},
probed for correspondence~\cite{dift}, and fine-tuned for amodal completion~\cite{pix2gestalt,dvas,taco}. We repurpose a
predictive model instead. Its pre-training task is close to ours, so fine-tuning its predictor with a small decoder and LoRA~\cite{lora} suffices, and inference takes two deterministic passes.

\section{Method}
\label{sec:method}

\subsection{Problem setup}
\label{sec:setup}
A video contains frames $I_1,\dots,I_L$ and, for one target object, its binary visible (modal) masks
$m_1,\dots,m_L$. The task is to predict its amodal masks $a_t\supseteq m_t$, which also cover the
parts hidden by other objects. As in prior work~\cite{dvas,c2fseg,eoras}, the visible masks are given; in practice they
can come from any video object segmenter~\cite{xmem,cutie,sam2}.

\subsection{Background on V-JEPA}
\label{sec:prelim}
A V-JEPA world model~\cite{vjepa,vjepa2,vjepa21} has two transformers, an encoder $E$ that maps a clip $x$ to $N$
tokens, one per space-time tubelet, and a narrower predictor $P$. In pre-training, a random set $D$ of tubelets covering
large space-time blocks is masked, and the rest, $K$, is kept. The encoder reads only the kept tokens $x_K$; the
predictor receives their features and learnable mask tokens at the positions in $D$, and regresses the features that a
slowly updated target encoder $\bar E$ computes on the full clip:
\begin{equation}
  \min_{E,P}\; \big\| P\big(E(x_K), D\big) - \mathrm{sg}\big(\bar E(x)\big)_D \big\|_1 ,
  \label{eq:vjepa}
\end{equation}
where $\mathrm{sg}$ stops gradients; V-JEPA~2.1 adds the same loss on the context tokens, weighted by their distance to
the masked ones~\cite{vjepa21}. Because the target lives in representation space, the predictor need not render
unpredictable pixel detail; it learns what is likely to occupy a masked region given the visible context. We use the released ViT-G model, with a 48-block encoder (1.85B parameters), a 24-block predictor and features concatenated from four encoder depths~\cite{vjepa21}.

\subsection{Occlusion as masking}
\label{sec:oam}
\paragraph{Object-centric clip.} For frame $t$, let $(c_t,h_t,w_t)$ be the centre, height and width of the bounding box
of $m_t$. We crop a square of side
\begin{equation}
  s_t = \min\!\big(\max\!\big(2\max(h_t,w_t),\, s_{\min}\big),\, s_{\max}\big)
  \label{eq:box}
\end{equation}
centred at $c_t$ and shifted into the frame (into a padded canvas on TAO-Amodal; Supp.~\cref{sec:s_impl}), and resize
it to $256\times256$. In frames where the object is invisible, centre and side are interpolated from the other frames.
A clip $x$ holds $T{=}24$ such crops and the cropped visible masks $\tilde m$; with tubelets of $2\times16\times16$ pixels,
it has $N=12\times16\times16$ tokens.

\paragraph{Dropping what is not visible.} Let $V_i$ be the pixels of tubelet $i$ and
\begin{equation}
  \rho_i = \frac{1}{|V_i|}\sum_{p\in V_i}\tilde m(p)
  \label{eq:cover}
\end{equation}
its coverage by the visible mask. We keep $K=\{i:\rho_i\ge\tau\}$ and drop $D=\{i:\rho_i<\tau\}$, with $\tau=0.25$. The dropped set holds the hidden parts of the object but also the occluders and the background. No occluder mask is needed, since the model decides where the object continues.

\paragraph{Latent completion.} The encoder computes the features of the full clip,
$z=\mathrm{LN}\big(E(x)\big)\in\mathbb R^{N\times d}$ ($d$ channels), normalised per feature level as in pre-training, and the
context features $E(x_K)$ of the kept tokens only. The predictor then fills in the dropped positions:
\begin{equation}
  \hat z_D = P\big(E(x_K), D\big), \qquad
  f_i = \begin{cases} \hat z_i, & i \in D,\\ z_i, & i \in K. \end{cases}
  \label{eq:fill}
\end{equation}
This is the pre-training computation of \cref{eq:vjepa}, except that the masked set is defined by the object's visibility instead of random blocks. Each dropped token now has two descriptions, $z_i$ for what the clip shows there and $f_i$ for what the predictor infers from the object's visible part alone. Behind an occluder, $z_i$ describes the occluder and
$f_i$ can describe the object (\cref{sec:exp_abl}).

\paragraph{Why the pre-trained predictor should help.} In pre-training, the target at a masked position is what the clip
shows there, an occluder included; amodal completion asks what lies behind the occluder. Under the L1 objective of
\cref{eq:vjepa}, the optimal prediction for a masked token is the per-channel median of its target feature given the context. In
\cref{eq:fill}, the context holds only the object's visible tokens and little evidence of the occluder, so the best
estimate at a dropped position should be what typically lies next to such an object, that is, the object where its shape is likely to continue and background elsewhere. The pre-trained weights should therefore be a strong starting point for completion,
which a randomly initialised predictor must learn from the amodal labels alone (\cref{sec:exp_abl}).

\paragraph{Mask decoder.} The decoder $H$ maps the token grid of $[z;f]\in\mathbb R^{N\times2d}$ to amodal logits. A
LayerNorm and a linear layer project each token to 256 channels; transposed convolutions upsample the
$12\times16\times16$ grid to $24\times128\times128$, and the visible mask is fused at the two finest scales. A
sub-pixel refiner (80K parameters) that also reads the RGB crop and $\tilde m$ raises the output to
$24\times256\times256$ logits $\ell$. The predicted amodal mask of frame $t$ is
\begin{equation}
  \hat a_t = \big(\Pi_t(\ell_t)>0\big)\,\lor\, m_t ,
  \label{eq:paste}
\end{equation}
where $\Pi_t$ resizes the crop-space logits and pastes them into the frame; longer videos use consecutive windows.

\paragraph{Training.} The predictor is initialised with its released weights and fine-tuned. The encoder stays frozen
except for LoRA~\cite{lora} adapters (rank 16) on its last 12 blocks, and the same adapted encoder computes $z$ and
$E(x_K)$. With $y=\sigma(\ell)$ and the cropped amodal target $\tilde a$, the loss is binary cross-entropy plus a Dice term:
\begin{equation}
  \mathcal L = \mathrm{BCE}(\ell,\tilde a) + 1-\frac{2\sum y\,\tilde a+1}{\sum y+\sum \tilde a+1}.
  \label{eq:loss}
\end{equation}
We train one model for 6000 steps of 8 clips, each batch drawn from SAIL-VOS~\cite{sailvos} or
MOVi-D~\cite{kubric,eoras} at random; MOVi-D lets the same weights serve synthetic benchmarks.

\subsection{Test-time re-crop}
\label{sec:recrop}
The crop of \cref{eq:box} is defined by the visible part. When most of the object is hidden, its amodal extent can exceed
the crop, and the part outside cannot be predicted. We therefore run a second pass, in which the boxes are recomputed with \cref{eq:box} from the
first-pass prediction $\hat a_t$ instead of $m_t$, and the same model predicts again on the new crops. Re-crop has no
parameters and doubles the inference cost; Algorithm 1 in Supp.~\cref{sec:s_impl} gives the full procedure.

\section{Experiments}
\label{sec:exp}

\begin{table*}[t]
\centering
\small
\begin{tabular*}{\textwidth}{@{\extracolsep{\fill}}llccccc@{}}
\toprule
& & \multicolumn{2}{c}{SAIL-VOS val} & \multicolumn{3}{c}{TAO-Amodal val} \\
\cmidrule(lr){3-4}\cmidrule(l){5-7}
Method & Type & mIoU & mIoU$_\text{occ}$ & AP$_{25}$ & AP$_{50}$ & AP$_{75}$ \\
\midrule
Visible mask & -- & 67.89 & n/a & 93.73 & 82.22 & 63.12 \\
Convex hull & -- & 63.18 & 27.54 & 93.73 & 82.22 & 63.12 \\
Convex hull (refined) & -- & 71.21 & 34.27 & 93.73 & 82.22 & 63.12 \\
PCNet-M~\cite{pcnet} & image & 74.20 & 42.52 & 94.89 & 85.11 & 65.97 \\
AISFormer~\cite{aisformer} & image & 73.51 & 39.16 & 95.45 & 81.93 & 59.84 \\
SDAmodal~\cite{sdamodal} & image & 72.73 & 41.26 & 94.43 & 83.60 & 63.06 \\
pix2gestalt~\cite{pix2gestalt} & image & 54.83 & 26.59 & 80.73 & 57.50 & 28.95 \\
VideoMAE~\cite{videomae} & video & 69.67 & 29.39 & 69.14 & 56.71 & 41.19 \\
3D-UNet~\cite{dvas} & video & 72.79 & 39.54 & 94.59 & 83.83 & 64.33 \\
Diffusion-VAS~\cite{dvas} & video & \underline{77.07} & \textbf{55.12} & \underline{97.28} & \underline{89.25} & \underline{71.99} \\
\midrule
\rowcolor{oursbg}
\method (ours) & video & \textbf{78.60} & \underline{52.09} & \textbf{98.68} & \textbf{93.36} & \textbf{76.84} \\
\bottomrule
\end{tabular*}
\caption{\textbf{Comparison with the state of the art} under the Diffusion-VAS protocols, with ground-truth visible masks; TAO-Amodal is evaluated zero-shot. Rows other than ours are as reported by~\cite{dvas}; ours is scored with the official metric. On TAO-Amodal, the convex hulls have the same box as the visible mask. The official mIoU$_\text{occ}$ averages each method over its own subset of frames; \cref{tab:controlled} scores ours and Diffusion-VAS on the same frames. Best in \textbf{bold}, second best \underline{underlined}.}
\label{tab:main}
\end{table*}

\begin{table}[t]
\centering
\small
\setlength{\tabcolsep}{3pt}
\resizebox{\linewidth}{!}{%
\begin{tabular}{@{}lccccc@{}}
\toprule
 & SAIL-VOS & \multicolumn{2}{c}{TAO test} & \multicolumn{2}{c}{Cost per frame} \\
\cmidrule(lr){2-2}\cmidrule(lr){3-4}\cmidrule(l){5-6}
Method & mIoU$_\text{occ}^\text{all}$ & AP$_{50}$ & AP$_{75}$ & TFLOPs & Time (ms) \\
\midrule
Visible mask & 0.00 & 83.08 & 62.42 & -- & -- \\
Diffusion-VAS~\cite{dvas} & \underline{39.98} & \underline{88.99} & \underline{71.05} & 36.1 & 471 \\
\rowcolor{oursbg}
\method (ours) & \textbf{43.68} & \textbf{92.99} & \textbf{76.40} & \textbf{1.48} & \textbf{14.6} \\
\midrule
$\Delta$ vs.\ Diffusion-VAS & +3.70 & +4.00 & +5.35 & & \\
 & {\scriptsize [+1.38, +5.40]} & {\scriptsize [+2.99, +5.03]} & {\scriptsize [+4.15, +6.56]} & & \\
\bottomrule
\end{tabular}}
\caption{\textbf{Same-frame comparison and cost}, all scored by us. mIoU$_\text{occ}^\text{all}$ averages the occluded IoU over all \nSvNset occluded SAIL-VOS frames, with misses counted as 0. TAO test is our held-out split; $\Delta$ is the difference with its 95\% CI over videos. Cost per object and frame on one GPU, for 24-frame windows (two passes) of \method and 25-frame clips (depth + 25 denoising steps) of Diffusion-VAS.}
\label{tab:controlled}
\end{table}

\begin{table}[t]
\centering
\small
\setlength{\tabcolsep}{3pt}
\resizebox{\linewidth}{!}{%
\begin{tabular}{@{}lcccc@{}}
\toprule
Method & all & out of frame & in-frame occl. & vis $<$ 0.5 \\
\midrule
\multicolumn{5}{@{}l}{\textit{\footnotesize val}} \\
Diffusion-VAS~\cite{dvas} & 89.25 & 82.85 & 87.66 & 37.12 \\
\rowcolor{oursbg}
\method & \textbf{93.36} & \textbf{87.39} & \textbf{93.33} & \textbf{65.68} \\
\midrule
\multicolumn{5}{@{}l}{\textit{\footnotesize test (held out)}} \\
Diffusion-VAS~\cite{dvas} & 88.99 & 80.25 & 87.86 & 43.75 \\
\rowcolor{oursbg}
\method & \textbf{92.99} & \textbf{85.61} & \textbf{94.27} & \textbf{73.19} \\
\midrule
\multicolumn{5}{@{}l}{\textit{\footnotesize test, visible masks tracked by SAM2}} \\
Diffusion-VAS~\cite{dvas} & 81.27 & 77.49 & 75.52 & 28.86 \\
\rowcolor{oursbg}
\method & \textbf{85.59} & \textbf{82.66} & \textbf{82.45} & \textbf{51.07} \\
\bottomrule
\end{tabular}}
\caption{\textbf{TAO-Amodal AP$_{50}$ by frame type.} Out of frame covers frames whose amodal box extends beyond the image, and in-frame occl.\ those with the box inside the image and the object partly hidden, of which vis $<$ 0.5 keeps those with less than half of the object visible. Bottom rows use visible masks tracked by SAM2 instead of annotated ones.}
\label{tab:taotype}
\end{table}

\begin{figure*}[t]
  \centering
  \includegraphics{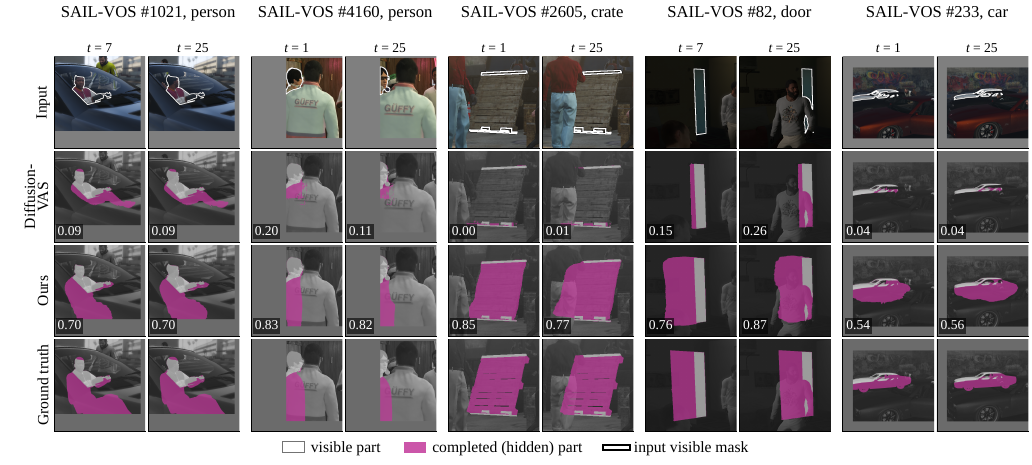}\\[2pt]
  \includegraphics{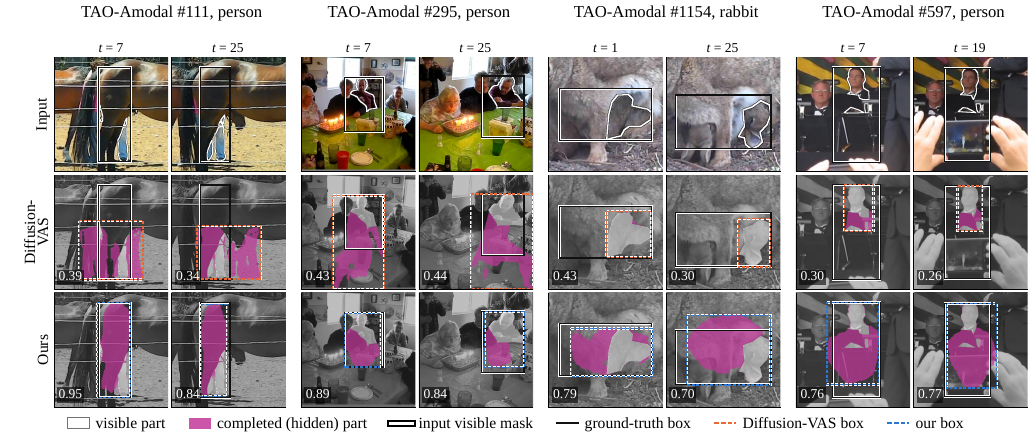}
  \caption{\textbf{Qualitative results}, two distant frames per sequence. Numbers give the occluded IoU on SAIL-VOS (top) and the box IoU on TAO-Amodal (bottom), which annotates only amodal boxes. Magenta marks the completed hidden part, and grey the area outside the image.}
  \label{fig:qual}
\end{figure*}

\begin{figure*}[t]
  \centering
  \includegraphics[width=\textwidth]{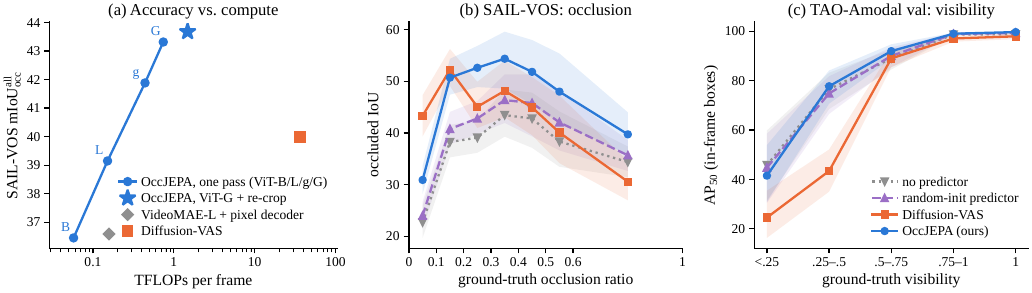}
  \caption{\textbf{Analysis.} (a) SAIL-VOS mIoU$_\text{occ}^\text{all}$ against inference FLOPs per frame for four V-JEPA
  world models (one pass), the final model with re-crop, VideoMAE-L and Diffusion-VAS. (b) SAIL-VOS occluded IoU
  by occlusion ratio. (c) TAO-Amodal val AP$_{50}$ by visibility, for amodal boxes inside the image. Bands show 95\% CIs over videos.}
  \label{fig:analysis}
\end{figure*}

We ask whether \method is more accurate than the diffusion-based state of the art on real and synthetic videos (\cref{sec:exp_sota,sec:exp_synth}), where its gain comes from (\cref{sec:exp_abl}) and what it costs (\cref{sec:exp_cost}).

\subsection{Setup}
\label{sec:exp_setup}

\paragraph{Benchmarks and metrics.} We follow the protocols of Diffusion-VAS~\cite{dvas}. \textbf{SAIL-VOS}~\cite{sailvos}
val has \nSvNseq sequences of 25 frames from \nSvNvid game-engine videos and is scored with the official amodal mIoU and
occluded-region mIoU$_\text{occ}$. The official mIoU$_\text{occ}$ averages each method only over the frames on which its
occluded IoU is non-zero, so the frames it misses do not count; Diffusion-VAS misses \nDvasMissedPct of the occluded
frames. We therefore also report mIoU$_\text{occ}^\text{all}$, the same per-frame
score averaged over all \nSvNset occluded frames, with misses counted as zero. \textbf{TAO-Amodal}~\cite{taoamodal}
annotates amodal boxes in real videos and is scored by AP$_{25}$, AP$_{50}$ and AP$_{75}$, the share of frames whose
amodal box has an IoU above 0.25, 0.5 and 0.75 with the ground truth. Besides val (\nTaoNframes frames), we evaluate on
a held-out \textbf{test split} (\nTtNframes frames, \nTtNvid videos) built with our reconstruction of the Diffusion-VAS
sequence rule (Supp.~\cref{sec:s_eval}). We also
evaluate on the synthetic \textbf{MOVi-D}, \textbf{MOVi-B}~\cite{kubric,eoras}, \textbf{Kubric-Static} and
\textbf{Kubric-Dynamic}~\cite{taco} benchmarks.

\paragraph{Baselines and implementation.} Diffusion-VAS is the state of the art on SAIL-VOS and TAO-Amodal. We score
its released outputs on SAIL-VOS and TAO-Amodal val, and run its released checkpoint on our test split and on Kubric;
on val, our run reproduces the released outputs (Supp.~\cref{sec:s_eval}). Unless stated otherwise, all methods
receive the ground-truth visible masks, and we report differences with 95\% bootstrap confidence intervals (CIs) over
videos.

\subsection{Comparison with the state of the art}
\label{sec:exp_sota}

\paragraph{SAIL-VOS and TAO-Amodal.} \Cref{tab:main} compares \method with every method reported by Diffusion-VAS.
\method has the best mIoU on SAIL-VOS (\nSvMiouOurs) and the best AP on TAO-Amodal at every threshold. It is second
on the official mIoU$_\text{occ}$ (\nSvOffOccOurs vs.\ \nSvOffOccDvas), which averages each method over its own frames
(\cref{sec:exp_setup}); scored on the same frames (\cref{tab:controlled}), it completes the occluded region better
(\nSvOccDiff mIoU$_\text{occ}^\text{all}$, CI \nSvOccDiffCI). On TAO-Amodal, it improves AP$_{50}$ on val (\nTaoAPDiff)
and on the held-out test split (\nTtAPDiff), so the gain holds on a split not used during development. A single
prediction of \method even beats the best of three Diffusion-VAS samples selected with the ground truth in AP$_{25}$
and AP$_{50}$ (Supp.~\cref{tab:s_best3}).

\paragraph{Qualitative results.} On SAIL-VOS (\cref{fig:qual}, top), Diffusion-VAS completes little of the hidden
objects, while \method recovers most of their shape, including that of a door behind a passing person. On
TAO-Amodal (bottom), \method recovers the extent of people and of a rabbit, whereas the boxes of Diffusion-VAS stay
near the visible part or spill onto neighbouring objects.

\paragraph{Occlusion level.} The advantage of \method is largest under heavy occlusion. On TAO-Amodal test frames with less than
half of the object visible, the AP$_{50}$ difference reaches \nTtLowVisDiff, and \method leads both when the object
leaves the image and when it is occluded inside it (\cref{tab:taotype}). On val, our variants without the pre-trained
predictor are about as far ahead at such low visibility (\cref{fig:analysis}c), so on TAO-Amodal this advantage comes
mainly from their shared crop, encoder and decoder. On SAIL-VOS (\cref{fig:analysis}b), the predictor adds accuracy at
every occlusion level, and \method leads Diffusion-VAS in every bin from 20\% occlusion upwards, by \nOcSvGapMin to
\nOcSvGapMax points. Below 20\%, the hidden part is often a token-sized sliver, which pixel-level generation fits better.

\paragraph{Predicted visible masks.} In practice, a video object segmenter supplies the visible masks. We track
them on the test split with SAM2~\cite{sam2} from the annotated first frame (mean IoU \nSamIoU with the annotations) and
run both methods unchanged. \method still leads in AP$_{50}$ (\nSamAPOurs vs.\ \nSamAPDvas;
\nSamAPDiff, CI \nSamAPDiffCI), and the gap grows to \nSamLowVisDiff when less than half of the object is
visible (\cref{tab:taotype}, bottom).

\subsection{Synthetic benchmarks}
\label{sec:exp_synth}

\begin{table}[t]
\centering
\small
\setlength{\tabcolsep}{2.5pt}
\resizebox{\linewidth}{!}{%
\begin{tabular}{@{}lcccccc@{}}
\toprule
 & \multicolumn{2}{c}{MOVi-D} & \multicolumn{2}{c}{MOVi-B$^\ast$} & \multicolumn{2}{c}{Kubric, mIoU} \\
\cmidrule(lr){2-3}\cmidrule(lr){4-5}\cmidrule(l){6-7}
Method & mIoU & mIoU$_\text{occ}$ & mIoU & mIoU$_\text{occ}$ & Static & Dynamic \\
\midrule
Diffusion-VAS~\cite{dvas} & \underline{82.97} & 47.86 & \underline{87.80} & \underline{53.69} & \underline{92.1} & 89.7 \\
\method, no predictor & 81.94 & \underline{48.17} & 86.02 & 52.62 & 91.4 & \underline{91.2} \\
\rowcolor{oursbg}
\method & \textbf{84.66} & \textbf{52.00} & \textbf{88.94} & \textbf{57.00} & \textbf{93.3} & \textbf{93.2} \\
\bottomrule
\end{tabular}}
\caption{\textbf{Synthetic benchmarks}, whole-frame input. On MOVi, Diffusion-VAS is as reported; on Kubric, we run both methods with the same inputs and scorer. $^\ast$Zero-shot for \method; Diffusion-VAS is trained on MOVi-B. Under the cropped EoRaS protocol, which fixes the input crop, Diffusion-VAS is ahead on MOVi (Supp.~\cref{tab:s_movi_crop}, with prior methods). Best in \textbf{bold}, second best \underline{underlined}.}
\label{tab:synth}
\end{table}

\begin{figure}[t]
  \centering
  \includegraphics{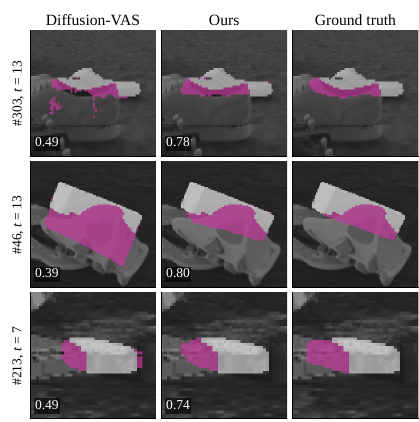}
  \caption{\textbf{Kubric-Dynamic.} Numbers give the occluded IoU, and magenta marks the completed hidden part.}
  \label{fig:kubric}
\end{figure}

On Kubric (\cref{tab:synth}; same inputs and scorer for both models), \method leads Diffusion-VAS in mIoU on Kubric-Static (\nKubSDiff) and Kubric-Dynamic (\nKubDDiff; \nKubDOccDiff on the hidden region
alone). The margin is larger on Kubric-Dynamic, where objects move and the occluded region changes shape; there, the
hidden part predicted by Diffusion-VAS breaks into fragments (\cref{fig:kubric}, top) or spills past the object's
outline (middle), while \method follows the ground-truth shape. On MOVi, where every method sees the whole frame,
\method leads on MOVi-D (\nMovidOurs vs.\ \nMovidDvas mIoU) and on MOVi-B (\nMovibOurs vs.\ \nMovibDvas), on which
Diffusion-VAS is trained and \method is not; on the hidden region, the margins are \nMovidOccDiff and \nMovibOccDiff
mIoU$_\text{occ}$. The predictor improves every column over the same model without it. All results come from one set
of weights, whereas Diffusion-VAS uses a separate model for MOVi-B.

\begin{table}[!tb]
\centering
\small
\setlength{\tabcolsep}{3pt}
\resizebox{\linewidth}{!}{%
\begin{tabular}{@{}lccccc@{}}
\toprule
 & \multicolumn{3}{c}{SAIL-VOS} & \multicolumn{2}{c}{TAO val} \\
\cmidrule(lr){2-4}\cmidrule(l){5-6}
Model & mIoU & mIoU$_\text{occ}$ & mIoU$_\text{occ}^\text{all}$ & AP$_{50}$ & AP$_{75}$ \\
\midrule
\rowcolor{oursbg}
\method (full) & 78.60 & 52.09 & 43.68 & 93.36 & 76.84 \\
$-$ re-crop & 78.73 & 51.52 & 43.32 & 92.05 & 75.63 \\
$-$ pre-trained weights & 76.60 & 45.09 & 37.30 & 90.17 & 70.32 \\
$-$ predictor & 75.18 & 42.26 & 35.15 & 89.43 & 68.18 \\
\bottomrule
\end{tabular}}
\caption{\textbf{Ablation} (ViT-G; data, steps, decoder and LoRA fixed; learning rates in Supp.~\cref{sec:s_impl}). Each row removes one component from the full model; without pre-trained weights, the predictor starts from V-JEPA's random initialisation. All drops in the last two rows, and the drops of re-crop in mIoU$_\text{occ}^\text{all}$ and AP$_{50}$, have 95\% CIs below zero (\cref{sec:exp_abl}; Supp.~\cref{tab:s_ablation}).}
\label{tab:ablation}
\end{table}

\begin{figure*}[t]
  \centering
  \includegraphics{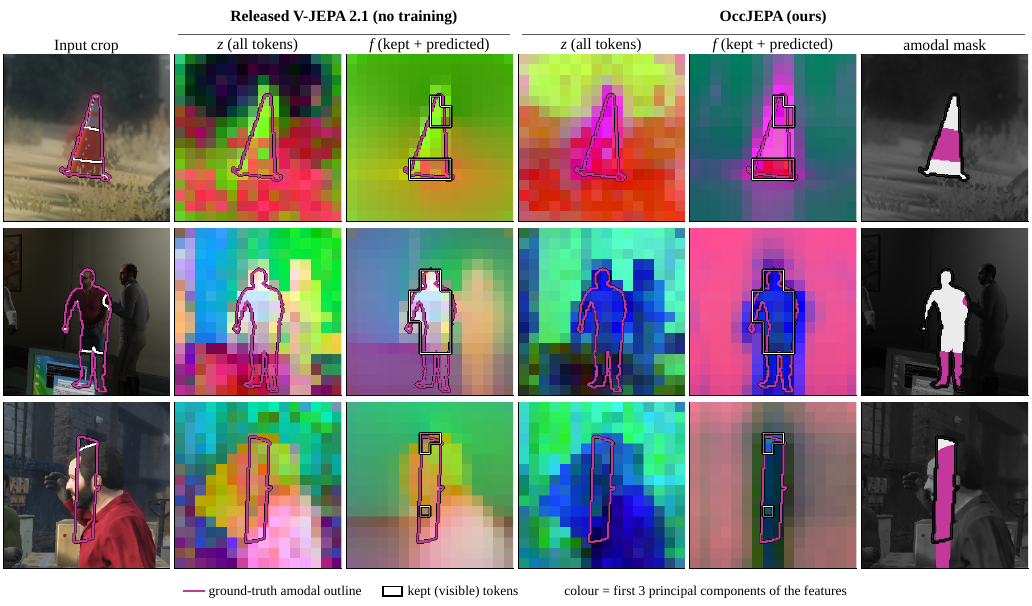}
  \caption{\textbf{What the predictor fills in.} Features as colours (first three principal components, one
  projection per model and example). $z$ (all tokens) shows the occluder where the object is hidden. In $f$ (kept tokens
  in black boxes, predicted elsewhere), the fine-tuned predictor of \method continues the object along its ground-truth
  outline, which the released predictor does only weakly.}
  \label{fig:features}
\end{figure*}

\begin{figure}[t]
  \centering
  \includegraphics{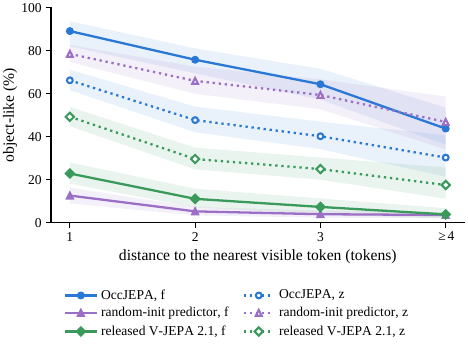}
  \caption{\textbf{Where the object is completed.} Share of hidden-object tokens (dropped, amodal coverage
  $\geq$0.5) whose feature is closer to the object's visible tokens than to the dropped background, by distance to the
  nearest visible token; $f$ solid, $z$ dotted; \nIbSeqs randomly sampled SAIL-VOS val sequences; 95\% CIs over videos.}
  \label{fig:interp}
\end{figure}

\subsection{Where does the gain come from?}
\label{sec:exp_abl}

The predictor puts \method ahead of Diffusion-VAS, mostly through its pre-trained weights, which shift the completion
from the encoder to the predictor.

\paragraph{The pre-trained predictor.} In \cref{tab:ablation}, which removes one component at a time, the predictor
adds \nSvPredAll mIoU$_\text{occ}^\text{all}$ on SAIL-VOS and \nTaoPredAll AP$_{50}$ on TAO-Amodal val (\nTtPred on the
test split). The same predictor trained from a random
initialisation adds only \nSvPredCap and \nTaoPredCap, so the pre-trained weights account for \nShareSv of the gain on
SAIL-VOS and \nShareTao on TAO-Amodal. With the same crops, decoder and data but no predictor, the model trails Diffusion-VAS on SAIL-VOS (\nNoneSvOcc vs.\ \nSvOccDvas
mIoU$_\text{occ}^\text{all}$) and is on par with it on TAO-Amodal (\nNoneTao vs.\ \nTaoAPDvas AP$_{50}$).

\paragraph{Where the object is completed.} In \cref{fig:features}, $z$ shows the occluder at hidden positions, while
$f$ continues the object behind it. \Cref{fig:interp} quantifies this. A hidden-object token counts as object-like when
its feature is closer to the object's visible tokens than to the dropped background tokens. In \method, $f$ is
object-like at \nIbOurF of the hidden-object tokens (\nIbOurZ for $z$) but at only \nIbOurBg of the background tokens, so
the predictor extends the object without spreading it into the background; the share falls with the distance to the
visible part. The
released predictor alone barely completes the object (\nIbRelF for $f$). Fine-tuned from the pre-trained weights, the
predictor reaches \nIbOurF; trained from a random initialisation, it reaches only \nIbRndF, and the adapted encoder
completes the object instead (\nIbRndZ for $z$), which gives a weaker model. The pre-trained weights thus shift the
completion from the encoder to the predictor.

\paragraph{Other world models and modules.} Across four V-JEPA~2.1 world models (\cref{fig:analysis}a;
Supp.~\cref{tab:s_wm}), the predictor improves over the same encoder alone at every size (\nGainG to \nGainB
mIoU$_\text{occ}^\text{all}$, one pass), and SAIL-VOS accuracy grows with size. Used in
place of the predictor, the pre-trained pixel decoder of VideoMAE-L~\cite{videomae} also adds \nMGain
mIoU$_\text{occ}^\text{all}$ over its encoder alone and \nMW over the same decoder randomly initialised, so the benefit of
a pre-trained completion module is not specific to JEPAs. At matched size (ViT-L, the V-JEPA model distilled from
ViT-G), latent prediction gives the stronger model, and VideoMAE-L scores \nMVsL mIoU$_\text{occ}^\text{all}$ and \nMVsLTao
AP$_{50}$ vs.\ V-JEPA.

\paragraph{Re-crop.} The second pass adds \nRcSv mIoU$_\text{occ}^\text{all}$ on SAIL-VOS and \nRcTao AP$_{50}$ on
TAO-Amodal val, and helps most where the first crop is too small, with \nRcLowVis AP$_{50}$ on test frames with visibility below
0.5 (CIs in Supp.~\cref{tab:s_ablation}).

\subsection{Cost}
\label{sec:exp_cost}

Diffusion-VAS runs a depth network on every frame and 25 denoising steps of a \nParamsDvasUnet-parameter video
diffusion U-Net. \method runs two deterministic forward passes of a \nParamsOurs-parameter model per 24-frame window.
Per frame of a long video, it needs \nCostOursFrameTF against \nCostDvasFrameTF TFLOPs (\nCostRatio$\times$ fewer) and, on the same NVIDIA RTX PRO 6000
Blackwell Max-Q GPU, \nSpOursMs against \nSpDvasMs ms (\nSpX$\times$ faster; \cref{tab:controlled}; other units in
Supp.~\cref{tab:s_cost}). Only \nTrainOurs of its parameters are trained, in \nTrainHours hours on one GPU with
\nTrainMem GB of memory. Smaller world models lower the cost further (\cref{fig:analysis}a). With the distilled ViT-L in
one pass (\nWmLFrameTF TFLOPs per frame, \nWmLRatio$\times$ fewer than Diffusion-VAS), \method already exceeds it on
TAO-Amodal val (\nWmLTaoDiff AP$_{50}$, CI \nWmLTaoDiffCI) and is within \nWmLGap of it on SAIL-VOS
mIoU$_\text{occ}^\text{all}$. On TAO-Amodal, even the 87M-parameter ViT-B beats it in one pass (\nWmBTao vs.\
\nTaoAPDvas AP$_{50}$).

\section{Conclusion}
\label{sec:conclusion}

The predictor of a self-supervised video world model is trained to infer what lies in masked regions, and amodal
segmentation asks a closely related question about occluded ones. Treating occlusion as masking turns the
V-JEPA~2.1 predictor into a completion module. The resulting model is deterministic, \nSpX$\times$ faster per frame
than Diffusion-VAS and more accurate on SAIL-VOS, TAO-Amodal and Kubric. The gain comes mainly from the predictor's
pre-trained weights, and SAIL-VOS accuracy grows with the size of the world model. The predictors of video world models are therefore worth keeping after pre-training, because they predict what a perception system cannot see.

\paragraph{Limitations.} Completion happens on $16\times16$-pixel tokens, so pixel-level generation does better when
only a thin sliver is hidden (\cref{fig:analysis}b). Without a temporal loss, \method is less temporally stable than
Diffusion-VAS (Supp.~\cref{tab:s_temporal}). Like prior work, it inherits the errors of its input visible masks
(\cref{tab:taotype}, bottom).

\par   %
\clearpage   %

{
    \small
    \bibliographystyle{ieeenat_fullname}
    \bibliography{refs}
}

\clearpage
\newcommand{\sRuleValSeqs}{1,426\xspace}
\newcommand{\sDvasValTracks}{1,389\xspace}
\newcommand{\sRuleBoth}{1,389\xspace}
\newcommand{\sRuleExtra}{37\xspace}
\newcommand{\sRuleSameWin}{1,370\xspace}
\newcommand{\sRuleSameWinPct}{98.6\%\xspace}
\newcommand{\sTestSeqs}{1,947\xspace}
\newcommand{\sTestVideos}{772\xspace}
\newcommand{\sPortFixedAP}{77.9\xspace}
\newcommand{\sPortFixedIoU}{0.972\xspace}
\newcommand{\sSvTrainSeqs}{15,688\xspace}
\newcommand{\sSvTrainDropped}{117\xspace}
\newcommand{\sSvTrainVideos}{155\xspace}
\newcommand{\sSvTrainInst}{15,223\xspace}
\newcommand{\sMoviTrainInst}{8,893\xspace}
\newcommand{\sPromoMiou}{80.95\xspace}
\newcommand{\sPromoOcc}{42.64\xspace}
\newcommand{\sPromoThrMiou}{80.17\xspace}
\newcommand{\sPromoThrOcc}{41.81\xspace}
\newcommand{\sPromoGMiou}{80.67\xspace}
\newcommand{\sPromoGOcc}{40.81\xspace}
\newcommand{\sRobFam}{+3.70\xspace}
\newcommand{\sRobFamCI}{[+0.78, +5.61]\xspace}
\newcommand{\sRobPre}{+3.70\xspace}
\newcommand{\sRobPreCI}{[+0.31, +5.70]\xspace}
\newcommand{\sRobSeq}{+3.70\xspace}
\newcommand{\sRobSeqCI}{[+2.85, +4.60]\xspace}
\newcommand{\sSeamFirstOurs}{7.99\xspace}
\newcommand{\sSeamRestOurs}{7.11\xspace}
\newcommand{\sSeamFirstDvas}{4.11\xspace}
\newcommand{\sSeamRestDvas}{3.96\xspace}
\newcommand{\sSeamFirstNone}{8.63\xspace}
\newcommand{\sSeamRestNone}{6.43\xspace}
\newcommand{\sSvMiouModal}{67.89\xspace}
\newcommand{\sSvOurDiff}{+7.89\xspace}
\newcommand{\sSvOurDiffCI}{[+6.09, +9.32]\xspace}
\newcommand{\sTaoLowValOurs}{65.68\xspace}
\newcommand{\sTaoLowValDvas}{37.12\xspace}
\newcommand{\sTaoOofValDiff}{+4.54\xspace}
\newcommand{\sTaoFullValOurs}{99.66\xspace}
\newcommand{\sTaoFullValDvas}{97.94\xspace}
\newcommand{\sTaoLowTestOurs}{73.19\xspace}
\newcommand{\sTaoLowTestDvas}{43.75\xspace}
\newcommand{\sTaoOofTestDiff}{+5.36\xspace}
\newcommand{\sTaoFullTestOurs}{98.81\xspace}
\newcommand{\sTaoFullTestDvas}{98.26\xspace}
\newcommand{\sAblRcMiou}{+3.42\xspace}
\newcommand{\sAblRcOcc}{+9.83\xspace}
\newcommand{\sAblRcApq}{+8.66\xspace}
\newcommand{\sWmTaoB}{+1.95\xspace}
\newcommand{\sWmTaoG}{+3.24\xspace}
\newcommand{\sWmTaoqB}{+5.57\xspace}
\newcommand{\sWmTaoqG}{+7.49\xspace}
\newcommand{\sGvsgTao}{+1.09\xspace}
\newcommand{\sMoviOccGainMin}{+1.86\xspace}
\newcommand{\sMoviOccGainMax}{+6.53\xspace}
\newcommand{\sKubDynOccDiff}{+7.71\xspace}
\newcommand{\sKubStaOccDiff}{+0.85\xspace}
\newcommand{\sKubDynOnePass}{51.78\xspace}
\newcommand{\sKubDynDvas}{44.33\xspace}
\newcommand{\sCompRcTao}{+1.24\xspace}
\newcommand{\sCompRcTaoNone}{$-$0.11\xspace}
\newcommand{\sCompHrPred}{+0.37\xspace}
\newcommand{\sCompHrNone}{+1.92\xspace}
\newcommand{\sCompMixSv}{+0.04\xspace}
\newcommand{\sCompMixTao}{$-$0.36\xspace}
\newcommand{\sDepthX}{9\xspace}
\newcommand{\sTrainPct}{3.6\%\xspace}
\newcommand{\sPredM}{59.4M\xspace}
\newcommand{\sHeadM}{3.9M\xspace}
\newcommand{\sLoraM}{5.7M\xspace}
\newcommand{\sTfB}{1.38\xspace}
\newcommand{\sOcSvShareHigh}{72\%\xspace}
\newcommand{\sOcSvNTop}{30,879\xspace}
\newcommand{\sOcTaoQMin}{41.5\xspace}
\newcommand{\sOcTaoHMin}{74.8\xspace}
\newcommand{\sQlRandOccMin}{70\%\xspace}
\newcommand{\sQlRandOccMax}{97\%\xspace}
\newcommand{\sQlRandDvasMax}{0.04\xspace}
\newcommand{\sQlRandOursMin}{0.29\xspace}
\newcommand{\sQlRandOursMax}{0.55\xspace}
\newcommand{\sQlWinOurs}{0.96\xspace}
\newcommand{\sQlWinDvas}{0.00\xspace}
\newcommand{\sQlAblSeqA}{2894\xspace}
\newcommand{\sQlAblOursA}{0.49\xspace}
\newcommand{\sQlAblRandA}{0.43\xspace}
\newcommand{\sQlAblNoneA}{0.43\xspace}
\newcommand{\sQlAblSeqB}{2742\xspace}
\newcommand{\sQlAblOursB}{0.26\xspace}
\newcommand{\sQlAblRandB}{0.06\xspace}
\newcommand{\sQlAblNoneB}{0.36\xspace}
\newcommand{\sQlAblSeqC}{4189\xspace}
\newcommand{\sQlAblOursC}{0.27\xspace}
\newcommand{\sQlAblRandC}{0.07\xspace}
\newcommand{\sQlAblNoneC}{0.01\xspace}
\newcommand{\sQlAblOccC}{90\%\xspace}
\newcommand{\sSamIoUMean}{0.826\xspace}
\newcommand{\sSamIoUMed}{0.919\xspace}
\newcommand{\sSamEmptyPct}{2.4\%\xspace}
\newcommand{\sSamDropOurs}{7.40\xspace}
\newcommand{\sSamDropDvas}{7.72\xspace}
\newcommand{\sSamDropModal}{7.43\xspace}
\newcommand{\sSamGainOurs}{+9.94\xspace}
\newcommand{\sGtGainOurs}{+9.91\xspace}
\newcommand{\sSamOnePassDiff}{+3.57\xspace}
\newcommand{\sSamOnePassDiffCI}{[+2.61, +4.56]\xspace}
\newcommand{\sSamRc}{+0.75\xspace}
\newcommand{\sSamLowVisDiffCI}{[+15.95, +28.09]\xspace}
\newcommand{\sSamLowOurs}{51.07\xspace}
\newcommand{\sSamLowDvas}{28.86\xspace}
\newcommand{\sBtDvasOneOcc}{55.12\xspace}
\newcommand{\sBtDvasThreeOcc}{59.69\xspace}
\newcommand{\sMissOursPct}{16.1\%\xspace}
\newcommand{\sMissDvasEmptyPct}{24.9\%\xspace}
\newcommand{\sMissDvasNoPixPct}{9.1\%\xspace}
\newcommand{\sMissDvasDisjointPct}{2.5\%\xspace}
\newcommand{\sTsPairDev}{+3.18\xspace}
\newcommand{\sTsPairDevCI}{[+1.99, +4.44]\xspace}
\newcommand{\sTsPairIoU}{$-$2.44\xspace}
\newcommand{\sTsPairIoUCI}{[$-$4.01, $-$0.92]\xspace}
\newcommand{\sTsPairDevNone}{+0.63\xspace}
\newcommand{\sTsPairDevNoneCI}{[+0.17, +1.12]\xspace}
\newcommand{\sDvasValSeqs}{1,392\xspace}
\newcommand{\sPortValIoU}{0.994\xspace}
\newcommand{\sPortValDiffQ}{$-$0.14\xspace}
\newcommand{\sPortValDiffQCI}{[$-$0.54, +0.27]\xspace}
\newcommand{\sPortValDiff}{$-$0.11\xspace}
\newcommand{\sPortValDiffCI}{[$-$0.80, +0.58]\xspace}
\newcommand{\sPortValDiffSev}{+0.06\xspace}
\newcommand{\sPortValDiffSevCI}{[$-$0.77, +0.89]\xspace}
\newcommand{\sPortValAP}{89.14\xspace}
\newcommand{\sPortOursDiff}{+4.22\xspace}
\newcommand{\sPortOursDiffCI}{[+3.13, +5.34]\xspace}

\setcounter{page}{1}
\makeatletter   %
\renewcommand{\thepage}{\texorpdfstring{\arabic{page}}{S\arabic{page}}}
\long\def\hyper@page@backref#1#2#3{\hyperlink{page.S#1}{#1}}   %
\def\@title{\thetitle\\[4pt]\large Supplementary Material}   %
\twocolumn[\supp@maketitle]\thispagestyle{plain}
\makeatother
\setcounter{section}{0}\renewcommand{\thesection}{\Alph{section}}\renewcommand{\theHsection}{S\Alph{section}}
\setcounter{table}{0}\renewcommand{\thetable}{S\arabic{table}}\renewcommand{\theHtable}{S\arabic{table}}
\setcounter{figure}{0}\renewcommand{\thefigure}{S\arabic{figure}}\renewcommand{\theHfigure}{S\arabic{figure}}
\renewcommand{\topfraction}{0.9}\renewcommand{\dbltopfraction}{0.9}\renewcommand{\textfraction}{0.08}
\renewcommand{\floatpagefraction}{0.75}\renewcommand{\dblfloatpagefraction}{0.75}
\setcounter{topnumber}{4}\setcounter{dbltopnumber}{3}\setcounter{totalnumber}{6}

This supplement gives implementation details (\cref{sec:s_impl}); the evaluation protocols (\cref{sec:s_eval}),
including how we build the TAO-Amodal test split, run Diffusion-VAS and track visible masks with SAM2; full results
with bootstrap confidence intervals (\cref{sec:s_results}); the cost, the analyses behind Fig.~\ref{fig:analysis} of
the main paper and the temporal behaviour (\cref{sec:s_analysis}); and further qualitative results
(\cref{sec:s_qual}).

\section{Implementation details}
\label{sec:s_impl}

\paragraph{Crops and windows.} For each frame, the crop is a square whose side is twice the longer side of the
visible-mask bounding box (Eq.~(\ref{eq:box}) of the main paper), clipped to $[s_{\min},s_{\max}]$. This range is 32 to 256 pixels on
MOVi and Kubric ($256{\times}256$ frames), 100 to 800 pixels on SAIL-VOS ($800{\times}1280$), and one eighth of the
shorter image side to the padded-canvas size on TAO-Amodal. On TAO-Amodal, the crop may extend beyond the image by a
quarter of the image height and width, onto the padded canvas on which Diffusion-VAS also predicts; elsewhere it is
shifted into the frame. In frames without visible pixels, centre and side are interpolated from the other frames. Each
crop is resampled to $256{\times}256$ with bilinear \texttt{grid\_sample}. Windows of 24 frames start at frames 1, 25,
49, \dots; when the length is not a multiple of 24, a last window ends at the last frame and overwrites the frames it
shares with the previous one. A 25-frame sequence is thus covered by frames 1--24 and 2--25. The logits are resized
bilinearly to the crop size, thresholded at zero, pasted into the frame and united with the visible mask
(Eq.~(\ref{eq:paste}) of the main paper).

\paragraph{Tokens and fill.} Tubelets of $2{\times}16{\times}16$ pixels give $12{\times}16{\times}16=3{,}072$ tokens per
24-frame clip. A token is kept if its visible-mask coverage $\rho_i$ (Eq.~(\ref{eq:cover}) of the main paper) is at
least $\tau=0.25$; if every token would be dropped, the most covered one is kept. The encoder returns four feature
levels, each normalised by its own layer norm, and the full-clip features $z$ are normalised per level by a
parameter-free layer norm, as in V-JEPA~2.1 pre-training. The mask decoder reads the concatenation $[z;f]$ of the
four-level full-clip and filled features.

\paragraph{Mask decoder.} The base decoder applies LayerNorm, a linear projection to 256 channels and GELU to every
token. Three transposed convolutions with GroupNorm and GELU then upsample the $16{\times}16$ grid to $128{\times}128$;
the visible mask is average-pooled and fused at $64{\times}64$ and $128{\times}128$, and a $1{\times}1$ convolution
outputs two frames per tubelet. The refiner takes, per frame, the RGB crop, the visible mask and the upsampled base
probabilities. It applies a pixel-unshuffle stem, three dilated convolutions (dilations 1, 2 and 4) that also receive
the last decoder features, and a zero-initialised convolution with pixel shuffle whose output is added to the logits
at $256{\times}256$.

\paragraph{Optimisation.} LoRA~\cite{S-lora} of rank 16 and scaling 1 adapts every linear layer (attention and MLP)
in the last quarter of the encoder blocks, which is the last 12 of 48 blocks for ViT-G and the last 3, 6 and 10 for ViT-B, ViT-L (also VideoMAE-L) and ViT-g. Its down-projection is initialised with Kaiming-uniform and its up-projection
with zeros. We use AdamW ($\beta=(0.9,0.999)$) with learning rates $10^{-3}$ for the mask decoder, $10^{-4}$ for the
pre-trained predictor ($10^{-3}$ for a randomly initialised one) and $5\cdot10^{-4}$ for LoRA, and weight decay 0.05 on
the weight matrices of the mask decoder and predictor (none on biases, norms, mask tokens and LoRA). The learning rates
warm up linearly for 300 steps and then decay to zero with a cosine schedule; gradients are clipped at 1.0. Training
runs 6000 steps of 8 clips, and we report the last checkpoint. The loss is Eq.~(\ref{eq:loss}) of the main paper,
averaged over the batch.

\paragraph{Data.} SAIL-VOS training uses the Diffusion-VAS training list without its \sSvTrainDropped sequences whose frames come from more than one video, which leaves \sSvTrainSeqs sequences from \sSvTrainVideos videos. Six training videos are held
out to monitor training. MOVi-D training uses the EoRaS training split (\sMoviTrainInst instances). Each batch comes
from SAIL-VOS or from MOVi-D with probability 0.5. As for Diffusion-VAS, the ground-truth visible masks are the input in
training and in every evaluation except the one with SAM2 masks (\cref{tab:s_sam2}).

\begin{figure}[t]
\fbox{\begin{minipage}{0.96\linewidth}\small
\textbf{Algorithm 1.\ Inference of \method.}\\[2pt]
\textbf{Input:} frames $I_{1:L}$, visible masks $m_{1:L}$.\quad \textbf{Output:} amodal masks $\hat a_{1:L}$.\\[2pt]
1:\ \ $u_t \leftarrow m_t$ for all $t$\\
2:\ \ \textbf{for} pass $=1,2$ \textbf{do}\\
3:\ \ \quad boxes $b_t \leftarrow$ Eq.~(\ref{eq:box}) on the bounding box of $u_t$\\
4:\ \ \quad \textbf{for} each window of $T$ frames \textbf{do}\\
5:\ \ \quad\quad crop $x,\tilde m$; coverage $\rho_i$ (\ref{eq:cover}); split tokens into $K,D$\\
6:\ \ \quad\quad $z\leftarrow\mathrm{LN}(E(x))$;\ \ $\hat z_D\leftarrow P(E(x_K),D)$;\ \ fill $f$ (\ref{eq:fill})\\
7:\ \ \quad\quad $\ell \leftarrow H([z;f],x,\tilde m)$;\ \ $\hat a_t \leftarrow (\Pi_t(\ell_t)>0)\lor m_t$ (\ref{eq:paste})\\
8:\ \ \quad $u_t \leftarrow \hat a_t$ for all $t$\\
9:\ \ \textbf{return} $\hat a_{1:L}$
\end{minipage}}
\end{figure}

\section{Evaluation protocols}
\label{sec:s_eval}

\paragraph{SAIL-VOS.} The official metric of Diffusion-VAS resizes each prediction to $800{\times}1280$, unites it with
the visible mask and computes its IoU with the amodal mask; for the occluded part, it computes the IoU of the predicted
and ground-truth occluded regions after a $5{\times}5$ erosion. The official mIoU$_\text{occ}$ averages this occluded
IoU over the frames where it is at least $10^{-6}$, so each method is averaged over the frames it does not miss
(``kept'' frames). Our mIoU$_\text{occ}^\text{all}$ averages the same per-frame value over all \nSvNset frames whose
eroded ground-truth occluded region is non-empty, counting misses as zero. On the released Diffusion-VAS outputs, our
re-implementation gives \nSvMiouDvas\,/\,\nSvOffOccDvas mIoU\,/\,mIoU$_\text{occ}$, against the reported
\nSvRepMiouDvas\,/\,\nSvRepOccDvas.

\paragraph{TAO-Amodal and the test split.} The metric turns each predicted amodal mask into a box, unites it with the
visible box, and reports the fraction of frames whose box IoU with the ground-truth amodal box exceeds 0.25, 0.5 and
0.75 (AP$_{25}$, AP$_{50}$, AP$_{75}$). Diffusion-VAS released the sequences of its TAO-Amodal val evaluation, but
neither the code that builds them nor any test sequences. We reconstructed the selection rule from the val release. It takes every BURST~\cite{S-burst} track in the videos at 30 frames per second and forms one sequence per track from the first 25 consecutive annotated frames that have both a BURST visible mask and a TAO-Amodal amodal box. On val, the rule
produces \sRuleValSeqs sequences. They contain all \sDvasValTracks tracks of the Diffusion-VAS release (plus
\sRuleExtra others), and for \sRuleSameWin of these tracks (\sRuleSameWinPct), the window and every field are
identical. The release itself has \sDvasValSeqs sequences (\nTaoNframes frames), since three of its tracks appear
twice. Applied to the test annotations, the rule gives \sTestSeqs sequences from \sTestVideos videos (\nTtNframes
frames).

\paragraph{Running Diffusion-VAS.} On the test split and Kubric, we run the official evaluation script with the
released checkpoint and all its settings (25 frames at $256{\times}512$, 25 denoising steps, Depth Anything~V2 ViT-L
pseudo-depth, fp16). Only the seed differs. The official script draws the noise for all sequences from one global
generator in an order that cannot be reproduced, so we seed each sequence with 23 plus its index. To check this
setup, we also run it on all \sDvasValSeqs val sequences (\nTaoNframes frames, \nTaoNvid videos), for which
Diffusion-VAS released its outputs. Our run reproduces the release. The median per-frame IoU between the two boxes is
\sPortValIoU, and the differences (run minus release) in AP$_{25}$, AP$_{50}$ and AP$_{75}$ are \sPortValDiffQ \sPortValDiffQCI,
\sPortValDiff \sPortValDiffCI and \sPortValDiffSev \sPortValDiffSevCI. \method leads our run by \sPortOursDiff
AP$_{50}$ \sPortOursDiffCI and the released outputs by \nTaoAPDiff.

\paragraph{Predicted visible masks.} SAM2.1 with the Hiera-L backbone~\cite{S-sam2} tracks each object of the test
split through its 25 frames, prompted with the annotated visible mask of the first frame, as in the standard
semi-supervised video object segmentation setting. The first frame keeps the prompt mask. On frames 2--25, the SAM2
masks have a mean IoU of \sSamIoUMean (median \sSamIoUMed) with the annotated masks; on the \sSamEmptyPct of these
frames where SAM2 returns an empty mask, we keep the last non-empty one. For every method, the tracked masks replace
the annotated ones, both as input and in the metric's union with the visible box; both methods run unchanged
(Diffusion-VAS with the same per-sequence seeds as above).

\paragraph{MOVi and Kubric.} On MOVi-D and MOVi-B, we use the EoRaS test lists. The primary counting rule, E (that of
EoRaS), averages over the frames with an annotated box; \cref{tab:s_movi} adds two other rules. In the cropped setting,
the input outside the EoRaS crop region is zeroed and predictions are clipped to it; in the whole-frame setting, the
input is the whole frame and our model computes its own crop, with re-crop. Kubric-Static and Kubric-Dynamic are the
TACO~\cite{S-taco} test sets at $256{\times}256$. TACO has not released its scorer, so we re-implemented it and use it
for every row we compute.

\paragraph{Statistics.} All intervals are 95\% bootstrap intervals over videos with 2000 resamples, paired for
differences; \cref{tab:s_sailvos} also resamples other units. Differences with an interval are computed before
rounding, so they can differ by 0.01 from the difference of two rounded entries.

\FloatBarrier   %
\section{Additional results}
\label{sec:s_results}

\begin{table*}[t]
\centering
\small
\setlength{\tabcolsep}{3pt}
\begin{tabular}{@{}lccccc@{}}
\toprule
Method & mIoU & mIoU$_\text{occ}$ & kept \% & mIoU$_\text{occ}^\text{all}$ & our$_\text{occ}$ \\
\midrule
Visible mask & 67.89 & n/a & 0.0 & 0.00 & 0.00 \\
Convex hull & 63.22 & 27.54 & 46.2 & 20.75 & 24.80 \\
Diffusion-VAS & 77.08 & 55.12 & 44.5 & 39.98 & 36.05 \\
\method & 78.60 & 52.09 & 51.4 & 43.68 & 43.94 \\
\midrule
$\Delta$ ours $-$ Diffusion-VAS & +1.52 {\scriptsize [+0.73, +2.33]} & $-$3.03 {\scriptsize [$-$5.26, $-$0.74]} &  & +3.70 {\scriptsize [+1.38, +5.40]} & +7.89 {\scriptsize [+6.09, +9.32]} \\
\bottomrule
\end{tabular}
\caption{\textbf{SAIL-VOS val, all metrics}, every row scored by us (the values reported by Diffusion-VAS, Tab.~\ref{tab:main} of the main paper, differ by at most 0.04). Kept is the share of all val frames that enter the official mIoU$_\text{occ}$, and our$_\text{occ}$ is the occluded IoU without erosion on frames whose occluded area is at least 16 pixels and 5\% of the amodal area, with misses counted as zero. The 95\% CI of $\Delta$ mIoU$_\text{occ}^\text{all}$ is \sRobSeqCI when the bootstrap resamples sequences, \sRobFamCI when it resamples the 23 game missions and \sRobPreCI when it resamples the 15 video-name prefixes.}
\label{tab:s_sailvos}
\end{table*}

\begin{table*}[t]
\centering
\small
\setlength{\tabcolsep}{3pt}
\begin{tabular}{@{}lcccccccccc@{}}
\toprule
 & \multicolumn{2}{c}{all} & \multicolumn{2}{c}{out of frame} & \multicolumn{2}{c}{in-frame occl.} & \multicolumn{2}{c}{vis $<$ 0.5} & \multicolumn{2}{c}{fully visible} \\
\cmidrule(lr){2-3}\cmidrule(lr){4-5}\cmidrule(lr){6-7}\cmidrule(lr){8-9}\cmidrule(lr){10-11}
Method & AP$_{50}$ & AP$_{75}$ & AP$_{50}$ & AP$_{75}$ & AP$_{50}$ & AP$_{75}$ & AP$_{50}$ & AP$_{75}$ & AP$_{50}$ & AP$_{75}$ \\
\midrule
Diffusion-VAS (val) & 89.25 & 71.99 & 82.85 & 55.12 & 87.66 & 68.40 & 37.12 & 11.57 & 97.94 & 94.09 \\
\method (val) & 93.36 & 76.84 & 87.39 & 57.72 & 93.33 & 77.15 & 65.68 & 26.85 & 99.66 & 96.49 \\
Diffusion-VAS (test) & 88.99 & 71.05 & 80.25 & 53.54 & 87.86 & 65.58 & 43.75 & 19.58 & 98.26 & 92.45 \\
\method (test) & 92.99 & 76.40 & 85.61 & 57.35 & 94.27 & 75.20 & 73.19 & 31.64 & 98.81 & 95.43 \\
Diffusion-VAS (test, SAM2) & 81.27 & 61.26 & 77.49 & 50.27 & 75.52 & 51.12 & 28.86 & 9.96 & 89.96 & 80.64 \\
\method (test, SAM2) & 85.59 & 66.85 & 82.66 & 54.14 & 82.45 & 60.45 & 51.07 & 18.99 & 91.14 & 84.54 \\
\bottomrule
\end{tabular}
\caption{\textbf{TAO-Amodal by frame type} (val / test frames, 34,800 / 48,675 in all, 11,124 / 16,200 out of frame, 13,068 / 15,300 with in-frame occlusion, 1,754 / 2,048 with vis $<$ 0.5 and 10,608 / 17,175 fully visible). Out of frame covers frames whose ground-truth amodal box leaves the image, and vis $<$ 0.5 those with the box inside the image and less than half of the object visible. SAM2 rows use visible masks tracked by SAM2 instead of annotated ones (\cref{tab:s_sam2}).}
\label{tab:s_tao}
\end{table*}

\paragraph{SAIL-VOS (\cref{tab:s_sailvos}).} \method leads on every metric that scores all methods on the same frames, by \nSvMiouDiff \nSvMiouDiffCI in mIoU, \nSvOccDiff \nSvOccDiffCI in mIoU$_\text{occ}^\text{all}$ and \sSvOurDiff \sSvOurDiffCI in our$_\text{occ}$, which also counts thin occluded regions that the $5{\times}5$ erosion removes. The lower
bound of the mIoU$_\text{occ}^\text{all}$ margin stays above zero whether the bootstrap resamples sequences, the
\nSvFamN game missions or the 15 video-name prefixes, so the gain is not confined to a few scenes. On the official
mIoU$_\text{occ}$, Diffusion-VAS is ahead (\nSvOffOccDiff \nSvOffOccDiffCI), but this metric averages each method over
a different set of frames, those on which its occluded IoU is non-zero. These are \nSvKeepOurs of all frames for \method and \nSvKeepDvas for Diffusion-VAS. Diffusion-VAS misses \nDvasMissedPct of the \nSvNset occluded frames. On \sMissDvasEmptyPct, its
predicted hidden part is empty after the metric's erosion (on \sMissDvasNoPixPct, it predicts no hidden pixel at all),
and on \sMissDvasDisjointPct, it does not overlap the ground truth. \method misses \sMissOursPct.
mIoU$_\text{occ}^\text{all}$ instead scores both methods on the same frames and counts misses as zero. The two reference
rows set the scale of the task. The visible mask alone reaches \sSvMiouModal mIoU, and the convex hull of the visible
mask, a purely geometric completion, reaches \nSvOccConvex mIoU$_\text{occ}^\text{all}$, less than half of the
\nSvOccOurs of \method.

\paragraph{TAO-Amodal by frame type (\cref{tab:s_tao}).} \method is ahead of Diffusion-VAS on every slice and at both
thresholds, on val and on the held-out test split alike, and the margin is largest on the hardest frames. When less
than half of the object is visible, AP$_{50}$ rises from \sTaoLowValDvas to \sTaoLowValOurs on val and from
\sTaoLowTestDvas to \sTaoLowTestOurs on test (\nTtLowVisDiff \nTtLowVisDiffCI). When the object extends beyond the
image, the margin is \sTaoOofValDiff AP$_{50}$ on val and \sTaoOofTestDiff on test, because the crop may extend onto the padded canvas and the model completes the object there as well. On fully visible objects, both methods are close to the
ceiling, and \method stays at \sTaoFullValOurs (val) and \sTaoFullTestOurs (test), against \sTaoFullValDvas and \sTaoFullTestDvas, so completing hidden parts does not make it over-extend visible objects.

\begin{table*}[t]
\centering
\small
\setlength{\tabcolsep}{3pt}
\begin{tabular}{@{}lcccccc@{}}
\toprule
 & \multicolumn{3}{c}{SAM2 visible masks} & \multicolumn{3}{c}{annotated visible masks} \\
\cmidrule(lr){2-4}\cmidrule(l){5-7}
Method & AP$_{25}$ & AP$_{50}$ & AP$_{75}$ & AP$_{25}$ & AP$_{50}$ & AP$_{75}$ \\
\midrule
Visible-mask box & 87.94 & 75.65 & 53.22 & 93.67 & 83.08 & 62.42 \\
Diffusion-VAS & 91.07 & 81.27 & 61.26 & 96.70 & 88.99 & 71.05 \\
\method, one pass & 92.01 & 84.84 & 66.29 & 97.53 & 92.10 & 76.09 \\
\method & 92.72 & 85.59 & 66.85 & 98.09 & 92.99 & 76.40 \\
\midrule
$\Delta$ \method{} $-$ Diffusion-VAS & +1.65 & +4.32 & +5.60 & +1.38 & +4.00 & +5.35 \\
\quad 95\% CI & {\scriptsize [+1.08, +2.26]} & {\scriptsize [+3.37, +5.34]} & {\scriptsize [+4.50, +6.76]} & {\scriptsize [+0.74, +2.01]} & {\scriptsize [+2.99, +5.03]} & {\scriptsize [+4.15, +6.56]} \\
$\Delta$ \method{} $-$ visible-mask box & +4.78 & +9.94 & +13.64 & +4.41 & +9.91 & +13.98 \\
\quad 95\% CI & {\scriptsize [+4.00, +5.65]} & {\scriptsize [+8.83, +11.21]} & {\scriptsize [+12.25, +15.04]} & {\scriptsize [+3.68, +5.23]} & {\scriptsize [+8.74, +11.16]} & {\scriptsize [+12.54, +15.49]} \\
\bottomrule
\end{tabular}
\caption{\textbf{TAO-Amodal test split with visible masks tracked by SAM2} (48,675 frames, 772 videos; \cref{sec:s_eval}). The same masks are the input of every method and enter the metric's union with the visible box. Differences come with 95\% CIs over videos.}
\label{tab:s_sam2}
\end{table*}

\paragraph{Predicted visible masks (\cref{tab:s_sam2}).} SAM2 masks lower the AP$_{50}$ of every method by about the same amount (\sSamDropModal for the visible-mask box itself, \sSamDropDvas for Diffusion-VAS and \sSamDropOurs for \method), so the drop comes from the tracked masks, not from the completion. The completion of \method adds
\sSamGainOurs AP$_{50}$ over the visible-mask box, as much as with annotated masks (\sGtGainOurs), and its lead over
Diffusion-VAS is \nSamAPDiff \nSamAPDiffCI, against \nTtAPDiff with annotated masks. A single pass of \method is
already ahead (\sSamOnePassDiff \sSamOnePassDiffCI), and re-crop adds \sSamRc. As with annotated masks, \method is ahead on every
slice of \cref{tab:s_tao} at both thresholds, and the AP$_{50}$ margin is largest when less than half of the object is visible, where \method reaches \sSamLowOurs AP$_{50}$ against \sSamLowDvas (\nSamLowVisDiff \sSamLowVisDiffCI).

\begin{table}[t]
\centering
\small
\setlength{\tabcolsep}{3pt}
\resizebox{\linewidth}{!}{%
\begin{tabular}{@{}lccccc@{}}
\toprule
 & \multicolumn{2}{c}{SAIL-VOS val} & \multicolumn{3}{c}{TAO-Amodal val} \\
\cmidrule(lr){2-3}\cmidrule(l){4-6}
Method & mIoU & mIoU$_\text{occ}$ & AP$_{25}$ & AP$_{50}$ & AP$_{75}$ \\
\midrule
pix2gestalt, one sample & 54.83 & 26.59 & 80.73 & 57.50 & 28.95 \\
pix2gestalt, best of 3 & 60.79 & 33.76 & 91.80 & 71.19 & 38.80 \\
Diffusion-VAS, one sample & 77.07 & 55.12 & 97.28 & 89.25 & 71.99 \\
Diffusion-VAS, best of 3 & 79.23 & 59.69 & 98.31 & 92.46 & 77.48 \\
\method, deterministic & 78.60 & 52.09 & 98.68 & 93.36 & 76.84 \\
\bottomrule
\end{tabular}}
\caption{\textbf{Best of three samples}, selecting for each frame the sample closest to the ground truth among three (an oracle). Rows other than ours are as reported by Diffusion-VAS.}
\label{tab:s_best3}
\end{table}

\paragraph{Best of three samples (\cref{tab:s_best3}).} For the two diffusion-based methods, pix2gestalt and
Diffusion-VAS, the Diffusion-VAS paper also reports the best of three samples, chosen for each frame with the ground truth. This
selection is an oracle, since it needs the answer at test time. It moves Diffusion-VAS from \sBtDvasOneOcc to
\sBtDvasThreeOcc mIoU$_\text{occ}$ on SAIL-VOS, a measure of how much its samples differ from one draw to the next.
\method is deterministic and gives a single answer. On TAO-Amodal, this answer exceeds the oracle best of three in
AP$_{25}$ and AP$_{50}$ but not in AP$_{75}$; on SAIL-VOS, the oracle puts Diffusion-VAS ahead.

\begin{table*}[t]
\centering
\small
\setlength{\tabcolsep}{3pt}
\begin{tabular}{@{}lccccc@{}}
\toprule
Difference (ViT-G) & mIoU & mIoU$_\text{occ}$ & mIoU$_\text{occ}^\text{all}$ & AP$_{50}$ & AP$_{75}$ \\
\midrule
pre-trained $-$ none & +2.92 {\scriptsize [+2.30, +3.65]} & +9.13 {\scriptsize [+7.51, +10.88]} & +8.19 {\scriptsize [+6.84, +9.72]} & +3.24 {\scriptsize [+2.50, +4.03]} & +7.49 {\scriptsize [+6.28, +8.83]} \\
random $-$ none & +1.00 {\scriptsize [+0.72, +1.29]} & +2.42 {\scriptsize [+1.89, +2.96]} & +1.81 {\scriptsize [+1.18, +2.43]} & +0.53 {\scriptsize [+0.17, +0.92]} & +1.70 {\scriptsize [+1.10, +2.32]} \\
pre-trained $-$ random & +1.92 {\scriptsize [+1.48, +2.45]} & +6.71 {\scriptsize [+5.39, +8.28]} & +6.38 {\scriptsize [+5.41, +7.57]} & +2.71 {\scriptsize [+2.12, +3.35]} & +5.79 {\scriptsize [+4.83, +6.84]} \\
pre-trained $-$ none, RC & +3.42 {\scriptsize [+2.77, +4.14]} & +9.83 {\scriptsize [+8.24, +11.64]} & +8.53 {\scriptsize [+7.16, +10.15]} & +3.93 {\scriptsize [+3.00, +4.84]} & +8.66 {\scriptsize [+7.39, +10.09]} \\
random $-$ none, RC & +1.42 {\scriptsize [+1.06, +1.79]} & +2.84 {\scriptsize [+2.20, +3.49]} & +2.15 {\scriptsize [+1.38, +2.86]} & +0.74 {\scriptsize [+0.34, +1.13]} & +2.14 {\scriptsize [+1.50, +2.82]} \\
pre-trained $-$ random, RC & +2.00 {\scriptsize [+1.50, +2.61]} & +6.99 {\scriptsize [+5.58, +8.90]} & +6.38 {\scriptsize [+5.38, +7.74]} & +3.19 {\scriptsize [+2.35, +4.01]} & +6.52 {\scriptsize [+5.47, +7.68]} \\
re-crop $-$ one pass & $-$0.13 {\scriptsize [$-$0.25, +0.02]} & +0.56 {\scriptsize [+0.26, +0.86]} & +0.36 {\scriptsize [+0.13, +0.58]} & +1.31 {\scriptsize [+0.68, +1.93]} & +1.21 {\scriptsize [+0.63, +1.78]} \\
\midrule
\multicolumn{6}{@{}l}{\textit{TAO-Amodal test}} \\
pre-trained $-$ none, RC & -- & -- & -- & +4.08 {\scriptsize [+3.43, +4.82]} & +9.38 {\scriptsize [+8.27, +10.59]} \\
re-crop $-$ one pass, vis $<$ 0.5 & -- & -- & -- & +11.33 {\scriptsize [+7.39, +15.22]} & +12.01 {\scriptsize [+8.32, +15.76]} \\
\bottomrule
\end{tabular}
\caption{\textbf{Completion-module ablation with 95\% CIs} (SAIL-VOS in the first three columns, TAO-Amodal val in the last two and the TAO-Amodal test split in the last block). RC marks re-crop, and vis $<$ 0.5 the frames whose amodal box is inside the image with less than half of the object visible.}
\label{tab:s_ablation}
\end{table*}

\begin{table}[t]
\centering
\small
\setlength{\tabcolsep}{3pt}
\resizebox{\linewidth}{!}{%
\begin{tabular}{@{}lcccccc@{}}
\toprule
 & & & \multicolumn{2}{c}{with module} & \multicolumn{2}{c}{gain} \\
\cmidrule(lr){4-5}\cmidrule(l){6-7}
Pre-trained model & Params (total) & TFLOPs & SV & AP$_{50}$ & SV & AP$_{50}$ \\
\midrule
\multicolumn{7}{@{}l}{\textit{pre-trained module: V-JEPA predictor, VideoMAE pixel decoder}} \\
V-JEPA 2.1 ViT-B$^\ast$ & 0.11B & 1.4 & 36.45 & 90.13 & +8.67 & +1.95 \\
V-JEPA 2.1 ViT-L$^\ast$ & 0.33B & 3.6 & 39.15 & 91.09 & +8.42 & +2.73 \\
V-JEPA 2.1 ViT-g & 1.08B & 10.6 & 41.88 & 90.96 & +8.45 & +2.93 \\
V-JEPA 2.1 ViT-G & 1.91B & 17.8 & 43.32 & 92.05 & +8.19 & +3.24 \\
VideoMAE-L & 0.35B & 3.8 & 36.58 & 90.25 & +9.83 & +2.51 \\
\midrule
\multicolumn{7}{@{}l}{\textit{same module, randomly initialised}} \\
V-JEPA 2.1 ViT-g & 1.08B & 10.6 & 36.16 & 88.69 & +2.72 & +0.66 \\
VideoMAE-L & 0.35B & 3.8 & 29.17 & 88.96 & +2.42 & +1.22 \\
\bottomrule
\end{tabular}}
\caption{\textbf{Completion modules of self-supervised video models}, same recipe (standard crop, one pass). SV is SAIL-VOS mIoU$_\text{occ}^\text{all}$, and AP$_{50}$ is measured on TAO-Amodal val. The gain is over the same encoder without any completion module, also for the randomly initialised modules. Parameters and TFLOPs (per 24-frame clip) are measured, and \cref{tab:s_worldmodels} gives the CIs of every difference. $^\ast$Distilled from ViT-G.}
\label{tab:s_wm}
\end{table}

\begin{table*}[t]
\centering
\small
\setlength{\tabcolsep}{3pt}
\begin{tabular}{@{}lcccc@{}}
\toprule
Difference (standard crop, one pass) & mIoU & mIoU$_\text{occ}^\text{all}$ & AP$_{50}$ & AP$_{75}$ \\
\midrule
V-JEPA ViT-B: predictor $-$ none & +2.80 {\scriptsize [+2.16, +3.65]} & +8.67 {\scriptsize [+7.41, +10.09]} & +1.95 {\scriptsize [+1.28, +2.62]} & +5.57 {\scriptsize [+4.48, +6.82]} \\
V-JEPA ViT-L: predictor $-$ none & +2.76 {\scriptsize [+2.00, +3.70]} & +8.42 {\scriptsize [+6.89, +10.18]} & +2.73 {\scriptsize [+2.05, +3.46]} & +6.32 {\scriptsize [+5.24, +7.50]} \\
V-JEPA ViT-g: predictor $-$ none & +2.70 {\scriptsize [+2.06, +3.49]} & +8.45 {\scriptsize [+7.29, +9.74]} & +2.93 {\scriptsize [+2.15, +3.75]} & +6.96 {\scriptsize [+5.82, +8.27]} \\
V-JEPA ViT-G: predictor $-$ none & +2.92 {\scriptsize [+2.30, +3.65]} & +8.19 {\scriptsize [+6.84, +9.72]} & +3.24 {\scriptsize [+2.50, +4.03]} & +7.49 {\scriptsize [+6.28, +8.83]} \\
VideoMAE-L: pixel decoder $-$ none & +3.40 {\scriptsize [+2.72, +4.22]} & +9.83 {\scriptsize [+8.59, +11.15]} & +2.51 {\scriptsize [+1.76, +3.27]} & +5.49 {\scriptsize [+4.47, +6.48]} \\
V-JEPA ViT-g: pre-trained $-$ random predictor & +1.58 {\scriptsize [+1.06, +2.15]} & +5.72 {\scriptsize [+4.83, +6.82]} & +2.28 {\scriptsize [+1.61, +2.93]} & +5.43 {\scriptsize [+4.50, +6.49]} \\
VideoMAE-L: pre-trained $-$ random pixel decoder & +2.50 {\scriptsize [+2.04, +3.04]} & +7.41 {\scriptsize [+6.45, +8.44]} & +1.29 {\scriptsize [+0.74, +1.85]} & +3.39 {\scriptsize [+2.55, +4.28]} \\
V-JEPA ViT-g: random predictor $-$ none & +1.12 {\scriptsize [+0.81, +1.44]} & +2.72 {\scriptsize [+2.03, +3.44]} & +0.66 {\scriptsize [+0.25, +1.08]} & +1.53 {\scriptsize [+0.92, +2.13]} \\
VideoMAE-L: random pixel decoder $-$ none & +0.90 {\scriptsize [+0.61, +1.24]} & +2.42 {\scriptsize [+1.82, +3.06]} & +1.22 {\scriptsize [+0.68, +1.78]} & +2.10 {\scriptsize [+1.40, +2.83]} \\
ViT-G $-$ ViT-g (with predictor) & +0.51 {\scriptsize [+0.20, +0.82]} & +1.43 {\scriptsize [+1.01, +1.90]} & +1.09 {\scriptsize [+0.72, +1.48]} & +1.40 {\scriptsize [+0.78, +2.04]} \\
VideoMAE-L $-$ V-JEPA ViT-L (with module) & $-$0.86 {\scriptsize [$-$1.26, $-$0.48]} & $-$2.57 {\scriptsize [$-$3.40, $-$1.73]} & $-$0.84 {\scriptsize [$-$1.31, $-$0.37]} & $-$0.81 {\scriptsize [$-$1.41, $-$0.20]} \\
\bottomrule
\end{tabular}
\caption{\textbf{World-model and module comparisons with 95\% CIs} (mIoU and mIoU$_\text{occ}^\text{all}$ on SAIL-VOS, AP on TAO-Amodal val).}
\label{tab:s_worldmodels}
\end{table*}

\begin{table*}[t]
\centering
\small
\setlength{\tabcolsep}{3pt}
\begin{tabular}{@{}lcccc@{}}
\toprule
Component added & SAIL-VOS mIoU$_\text{occ}^\text{all}$ & TAO AP$_{50}$ & TAO AP$_{75}$ & TAO out-of-frame AP$_{75}$ \\
\midrule
sub-pixel refiner, with predictor & +0.37 {\scriptsize [+0.12, +0.63]} & $-$0.11 {\scriptsize [$-$0.45, +0.28]} & $-$0.29 {\scriptsize [$-$0.86, +0.34]} & $-$2.02 {\scriptsize [$-$3.45, $-$0.66]} \\
sub-pixel refiner, no predictor & +1.92 {\scriptsize [+1.43, +2.42]} & $-$0.96 {\scriptsize [$-$1.44, $-$0.48]} & $-$0.97 {\scriptsize [$-$1.71, $-$0.25]} & $-$3.35 {\scriptsize [$-$5.22, $-$1.59]} \\
MOVi-D in training & +0.04 {\scriptsize [$-$0.30, +0.34]} & $-$0.36 {\scriptsize [$-$0.74, $-$0.01]} & +0.05 {\scriptsize [$-$0.51, +0.63]} & $-$0.78 {\scriptsize [$-$2.21, +0.66]} \\
re-crop, with predictor & +0.47 {\scriptsize [+0.26, +0.68]} & +1.24 {\scriptsize [+0.67, +1.84]} & +0.66 {\scriptsize [+0.16, +1.18]} & +1.13 {\scriptsize [$-$0.03, +2.38]} \\
re-crop, no predictor & +0.26 {\scriptsize [$-$0.03, +0.57]} & $-$0.11 {\scriptsize [$-$0.68, +0.53]} & $-$0.55 {\scriptsize [$-$1.09, +0.00]} & $-$0.53 {\scriptsize [$-$1.58, +0.55]} \\
\bottomrule
\end{tabular}
\caption{\textbf{Design components} (ViT-g world model; change from adding each component, with 95\% CIs; TAO-Amodal val). Out of frame covers the frames whose ground-truth amodal box leaves the image.}
\label{tab:s_components}
\end{table*}

\paragraph{Completion module (\cref{tab:s_ablation}).} Every predictor difference in the table has a 95\% CI above
zero, in one pass and with re-crop. The effect of the pre-trained predictor is concentrated where completion is needed. With
re-crop, it adds \sAblRcMiou to mIoU, which is dominated by the visible part, but \sAblRcOcc to mIoU$_\text{occ}$ and
\nSvPredAll to mIoU$_\text{occ}^\text{all}$. On TAO-Amodal, it adds more at the stricter threshold (\sAblRcApq
AP$_{75}$ against \nTaoPredAll AP$_{50}$), that is, it makes the amodal extent more precise. The same architecture
trained from a random initialisation recovers only \nSvPredCap mIoU$_\text{occ}^\text{all}$ and \nTaoPredCap AP$_{50}$
of these gains, so the pre-trained weights account for \nShareSv and \nShareTao of them. The decomposition is nearly
the same in one pass, so it does not depend on re-crop. Re-crop itself raises mIoU$_\text{occ}$,
mIoU$_\text{occ}^\text{all}$, AP$_{50}$ and AP$_{75}$, each with a CI above zero; its effect on mIoU, which the
visible part dominates, has a CI that includes zero.

\begin{figure*}[t]\centering
\includegraphics{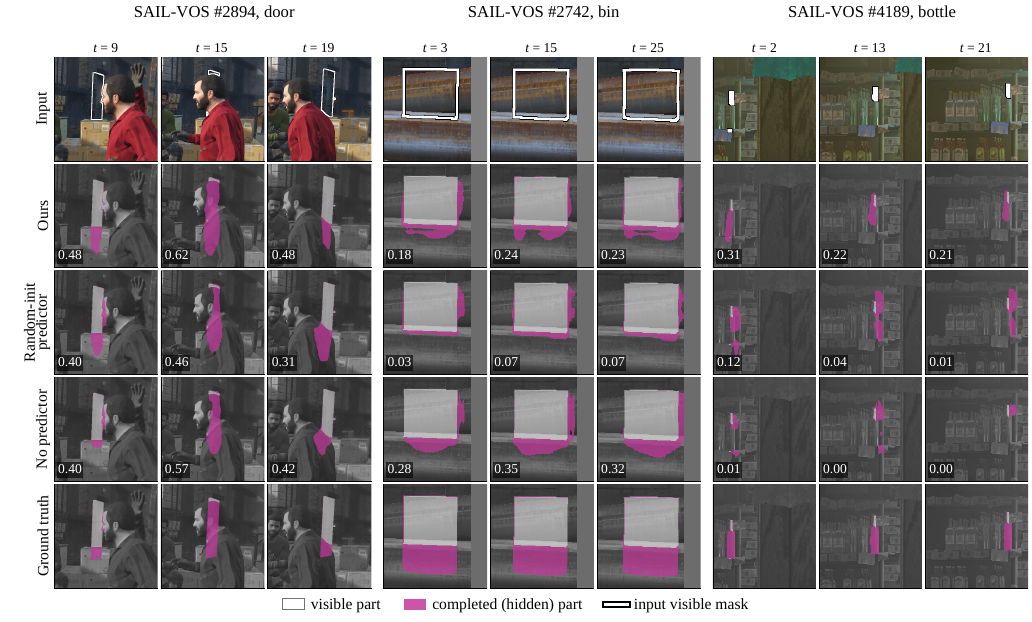}
\caption{\textbf{Predictor ablation on SAIL-VOS} with the pre-trained predictor (ours), the same predictor trained from a random initialisation and no predictor, next to the ground truth. From each third of the sequence, we show the frame whose gain of ours over the better ablation is closest to that gain over the whole sequence. Numbers give the per-frame occluded IoU.}
\label{fig:s_abl}
\end{figure*}

\begin{figure*}[t]\centering
\includegraphics{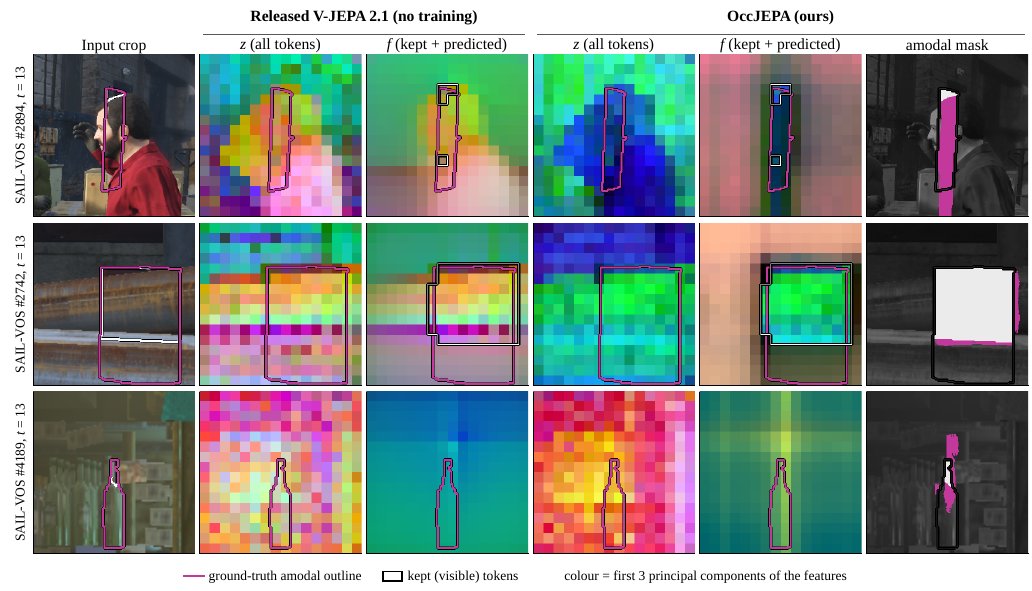}
\caption{\textbf{Feature visualisation on the sequences of \cref{fig:s_abl}}, middle frame. As in Fig.~\ref{fig:features} of the main paper, we show the first three principal components of the full-clip features $z$ and of the filled features $f$, for the released V-JEPA~2.1 and for \method, with the same legend.}
\label{fig:s_feat}
\end{figure*}

\paragraph{Predictor ablation, examples (\cref{fig:s_abl}).} Of these three sequences, the pre-trained predictor gives
the highest occluded IoU over the whole sequence in two. In sequence \sQlAblSeqA, it reaches \sQlAblOursA, against
\sQlAblRandA with a randomly initialised predictor and \sQlAblNoneA without one. In sequence \sQlAblSeqC, where \sQlAblOccC of the bottle is hidden, only
the pre-trained predictor recovers a substantial part of it (\sQlAblOursC vs.\ \sQlAblRandC and \sQlAblNoneC). In sequence \sQlAblSeqB, the model without a predictor covers more of the hidden lower part of the bin
(\sQlAblNoneB vs.\ \sQlAblOursB). Over the whole benchmark, the pre-trained predictor is ahead of both alternatives,
with every CI above zero (\cref{tab:s_ablation}).

\paragraph{Feature maps (\cref{fig:s_feat}).} These sequences show the same effect as Fig.~\ref{fig:features} of the
main paper. At hidden positions, $z$ describes what the frame shows there, the occluder or the background, while the
filled features $f$ of \method continue the object. In sequence \sQlAblSeqA, they form a narrow vertical region along
the hidden part of the door behind the man's head, and in sequence \sQlAblSeqC, a column along the bottle, of which only
a few tokens are visible. In sequence \sQlAblSeqB, the filled region stays close to the visible part of the bin, and
the mask decoder completes only part of the hidden region, as in \cref{fig:s_abl}.

\paragraph{World models and completion modules (\cref{tab:s_wm,tab:s_worldmodels}).} The distilled ViT-B and ViT-L
predictors output features in the space of their ViT-G teacher (1,664 channels), not in that of their own encoder (768
and 1,024 channels, one level). For these two models, the mask decoder therefore reads the encoder tokens $z$
concatenated with $f$, which holds the predictor output at dropped positions and zeros at kept ones; for ViT-g and
ViT-G, $z$ and $f$ share the four-level encoder space, as in the main paper. The predictor improves over the encoder
alone at every world-model size, by \nGainG to \nGainB mIoU$_\text{occ}^\text{all}$, each with a CI above zero. On
TAO-Amodal, its gain grows with the size of the world model, from \sWmTaoB AP$_{50}$ (ViT-B) to \sWmTaoG (ViT-G) and
from \sWmTaoqB to \sWmTaoqG AP$_{75}$, in line with the main-paper finding that SAIL-VOS accuracy grows with the world
model. With the predictor, moving from ViT-g to ViT-G adds \nGVsG mIoU$_\text{occ}^\text{all}$ and \sGvsgTao
AP$_{50}$. Pre-trained weights matter for both families of video models. Against the same module trained from scratch,
the V-JEPA ViT-g predictor gains \nGW and the VideoMAE-L pixel decoder \nMW mIoU$_\text{occ}^\text{all}$. VideoMAE-L gains
more over its own encoder (\nMGain), but from a weaker start, since its encoder alone reaches \nMEncSv against \nLEncSv for
V-JEPA ViT-L. With their completion modules, latent prediction (V-JEPA) gives the better model at matched size (ViT-L, with the V-JEPA model distilled from ViT-G) on both benchmarks. VideoMAE-L minus V-JEPA ViT-L is \nMVsL mIoU$_\text{occ}^\text{all}$ and \nMVsLTao AP$_{50}$, both with CIs below zero.

\begin{table*}[t]
\centering
\small
\setlength{\tabcolsep}{3pt}
\begin{tabular}{@{}llccccc@{}}
\toprule
Setting & Model & E: mIoU & E: mIoU$_\text{occ}$ & A: mIoU & A: mIoU$_\text{occ}$ & D: mIoU$_\text{occ}$ \\
\midrule
MOVi-D, cropped & no predictor & 74.99 & 39.85 & 63.46 & 33.72 & 46.90 \\
MOVi-D, cropped & ViT-g world model & 75.21 & 39.94 & 63.65 & 33.80 & 46.84 \\
MOVi-D, cropped & \method & 76.47 & 42.05 & 64.71 & 35.58 & 51.08 \\
\midrule
MOVi-D, whole frame & no predictor & 81.94 & 48.17 & 77.28 & 48.70 & 51.53 \\
MOVi-D, whole frame & ViT-g world model & 81.73 & 47.91 & 76.81 & 48.19 & 51.28 \\
MOVi-D, whole frame & \method & 84.66 & 52.00 & 79.79 & 52.16 & 57.36 \\
\midrule
MOVi-B, cropped & no predictor & 80.17 & 45.83 & 73.89 & 42.24 & 52.59 \\
MOVi-B, cropped & ViT-g world model & 80.39 & 46.08 & 74.10 & 42.47 & 52.96 \\
MOVi-B, cropped & \method & 81.87 & 48.70 & 75.45 & 44.88 & 57.46 \\
\midrule
MOVi-B, whole frame & no predictor & 86.02 & 52.62 & 83.91 & 53.13 & 57.85 \\
MOVi-B, whole frame & ViT-g world model & 86.12 & 52.85 & 83.81 & 53.14 & 58.12 \\
MOVi-B, whole frame & \method & 88.94 & 57.00 & 86.75 & 57.31 & 64.38 \\
\bottomrule
\end{tabular}
\caption{\textbf{MOVi under three frame-counting rules.} Rule E (primary) is that of EoRaS and averages over the frames with an annotated box. Rule A uses all frames with a non-empty amodal mask, including fully hidden ones, and rule D is that of Diffusion-VAS on SAIL-VOS (eroded occluded part, misses dropped). Whole-frame rows use re-crop.}
\label{tab:s_movi}
\end{table*}

\begin{table}[t]
\centering
\small
\setlength{\tabcolsep}{3pt}
\resizebox{\linewidth}{!}{%
\begin{tabular}{@{}lcccc@{}}
\toprule
 & \multicolumn{2}{c}{Kubric-Static} & \multicolumn{2}{c}{Kubric-Dynamic} \\
\cmidrule(lr){2-3}\cmidrule(l){4-5}
Model & mIoU & mIoU$_\text{occ}$ & mIoU & mIoU$_\text{occ}$ \\
\midrule
Diffusion-VAS (run by us) & 92.13 & 48.69 & 89.67 & 44.33 \\
no predictor, + re-crop & 91.38 & 44.83 & 91.19 & 45.73 \\
ViT-g world model, + re-crop & 92.03 & 47.54 & 91.92 & 49.53 \\
\method, one pass & 92.02 & 47.75 & 92.95 & 51.78 \\
\method, + re-crop & 93.31 & 49.54 & 93.23 & 52.04 \\
\bottomrule
\end{tabular}}
\caption{\textbf{Kubric with the occluded-region metric} (rule A, our re-implementation of the TACO scorer, $256{\times}256$).}
\label{tab:s_kubric}
\end{table}

\begin{table}[t]
\centering
\small
\setlength{\tabcolsep}{3pt}
\begin{tabular}{@{}lcc@{}}
\toprule
Method & Kubric-Static & Kubric-Dynamic \\
\midrule
pix2gestalt~\cite{S-pix2gestalt} & 75.6 & 68.3 \\
PCNet~\cite{S-pcnet} & 83.2 & 76.2 \\
TACO~\cite{S-taco} & 83.9 & 77.4 \\
Diffusion-VAS~\cite{S-dvas} & 84.3 & 77.8 \\
IC-Amodal~\cite{S-icamodal} & 86.7 & 85.5 \\
\bottomrule
\end{tabular}
\caption{\textbf{Kubric mIoU reported by~\cite{S-taco,S-icamodal}} with TACO's scorer. Our scorer gives Diffusion-VAS a different value (Tab.~\ref{tab:synth} of the main paper), so these values are compared only with each other.}
\label{tab:s_kubric_rep}
\end{table}

\begin{table}[t]
\centering
\small
\setlength{\tabcolsep}{3pt}
\resizebox{\linewidth}{!}{%
\begin{tabular}{@{}lcccc@{}}
\toprule
 & \multicolumn{2}{c}{MOVi-D} & \multicolumn{2}{c}{MOVi-B$^\ast$} \\
\cmidrule(lr){2-3}\cmidrule(l){4-5}
Method & mIoU & mIoU$_\text{occ}$ & mIoU & mIoU$_\text{occ}$ \\
\midrule
EoRaS~\cite{S-eoras} & -- & -- & 79.22 & 47.89 \\
EoRaS+SG~\cite{S-eoras} & 74.10 & 38.33 & 81.76 & 49.39 \\
C2F-Seg~\cite{S-c2fseg} & 71.67 & 36.13 & -- & -- \\
Amodal SAM~\cite{S-amodalsam} & 73.06 & 39.21 & -- & -- \\
Diffusion-VAS~\cite{S-dvas} & 77.03 & 44.23 & 83.51 & 53.75 \\
\method & 76.47 & 42.05 & 81.87 & 48.70 \\
\bottomrule
\end{tabular}}
\caption{\textbf{MOVi under the cropped EoRaS protocol}, with the input fixed to the EoRaS crop region. Other methods are as reported in their papers (rule E). $^\ast$\method is zero-shot on MOVi-B, whereas EoRaS and Diffusion-VAS are trained on it.}
\label{tab:s_movi_crop}
\end{table}

\paragraph{MOVi and Kubric (\cref{tab:s_movi,tab:s_kubric,tab:s_kubric_rep,tab:s_movi_crop}).} Under the cropped EoRaS
protocol (\cref{tab:s_movi_crop}), \method is ahead of EoRaS+SG, C2F-Seg and Amodal SAM on MOVi-D on both metrics.
Diffusion-VAS, trained on MOVi-B and MOVi-D, is ahead of \method under this protocol, which fixes the input crop and
thus disables our object-centric crop and re-crop; on MOVi-B, on which \method is zero-shot, EoRaS+SG also has a higher
mIoU$_\text{occ}$. Among our models, \method ranks first however frames are counted, with the best value in all five columns in each of the four settings of \cref{tab:s_movi}. On the occluded metrics, its margin over the same model
without a predictor ranges from \sMoviOccGainMin to \sMoviOccGainMax and is largest under rule D, that of
Diffusion-VAS on SAIL-VOS (eroded occluded region, misses dropped). The margins on MOVi-B, which none of our models saw
in training, are at least as large as on MOVi-D. On Kubric with the occluded-region metric (\cref{tab:s_kubric}),
\method with re-crop is the best model in all four columns. Its advantage over Diffusion-VAS is \sKubDynOccDiff
mIoU$_\text{occ}$ on Kubric-Dynamic, where a single pass already exceeds Diffusion-VAS (\sKubDynOnePass vs.\
\sKubDynDvas), and \sKubStaOccDiff on Kubric-Static.

\paragraph{Design components (\cref{tab:s_components}).} These ablations use the ViT-g world model. Re-crop recomputes
the crop from the first-pass amodal prediction, so it can only help when that prediction extends beyond the visible
part. Accordingly, the second pass adds \sCompRcTao AP$_{50}$ on TAO-Amodal with the predictor (CI above zero) and
\sCompRcTaoNone without it. The sub-pixel refiner sharpens the masks on SAIL-VOS (\sCompHrPred
mIoU$_\text{occ}^\text{all}$ with the predictor, \sCompHrNone without) but lowers TAO-Amodal AP$_{75}$ on boxes that
leave the image. Mixing MOVi-D into training leaves SAIL-VOS unchanged (\sCompMixSv mIoU$_\text{occ}^\text{all}$) and
TAO-Amodal within half a point (\sCompMixTao AP$_{50}$), and lets one model serve all benchmarks of the paper.

\section{Additional analyses}
\label{sec:s_analysis}

\begin{table}[t]
\centering
\small
\setlength{\tabcolsep}{3pt}
\resizebox{\linewidth}{!}{%
\begin{tabular}{@{}lcccccc@{}}
\toprule
Model & Total & Module & Mask dec. & LoRA & Trained & TFLOPs \\
\midrule
ViT-B$^\ast$ & 0.111 & 22.9 & 1.08 & 0.59 & 24.5 & 1.38 \\
ViT-L$^\ast$ & 0.330 & 23.0 & 1.14 & 1.57 & 25.7 & 3.64 \\
ViT-g & 1.076 & 55.4 & 3.36 & 3.77 & 62.5 & 10.57 \\
ViT-G (\method) & 1.914 & 59.4 & 3.88 & 5.70 & 69.0 & 17.78 \\
VideoMAE-L & 0.345 & 39.1 & 0.85 & 1.57 & 41.6 & 3.79 \\
\bottomrule
\end{tabular}}
\caption{\textbf{Cost of each model} (V-JEPA world models unless marked). Parameters are in billions for the total and in millions for the module (predictor or pixel decoder) and the trained part (module, LoRA and mask decoder). TFLOPs are per 24-frame $256{\times}256$ window and one pass (re-crop doubles them), counted with PyTorch's FLOP counter on the inference code. Diffusion-VAS has a 1.52B U-Net, a 0.098B VAE, a 0.63B image encoder and a 0.34B depth network, and needs 903 TFLOPs per 25-frame $256{\times}512$ clip. Per 25-frame benchmark sequence (two windows, frames 1--24 and 2--25), \method takes \nSpOursSeq\,s and \nCostOursSeqTF TFLOPs, Diffusion-VAS \nSpDvas\,s and \nCostDvasTF TFLOPs; per frame of a long video (windows at stride 24), \nSpOursMs\,ms and \nCostOursFrameTF TFLOPs against \nSpDvasMs\,ms and \nCostDvasFrameTF TFLOPs. Training \method takes \nTrainHours\,h on one GPU (\nTrainMem\,GB peak memory). $^\ast$Distilled from ViT-G.}
\label{tab:s_cost}
\end{table}

\paragraph{Cost (Tab.~\ref{tab:controlled} of the main paper and \cref{tab:s_cost}).} We time both methods on the same
NVIDIA RTX PRO 6000 Blackwell Max-Q GPU and the same 50 Kubric-Static clips, with batch size 1 and no warm-up, including pre- and post-processing but not
model loading (\method in bf16, Diffusion-VAS in fp16 as released). With re-crop, \method needs \nCostOursTF TFLOPs and
\nSpOurs\,s per object and 24-frame window; Diffusion-VAS needs \nCostDvasTF TFLOPs and \nSpDvas\,s for the 25-frame
clip that covers the same 24 frames (\nCostRatioClip$\times$ and \nSpXClip$\times$ more). Per frame of a long video,
where windows do not overlap, the ratios are \nCostRatio$\times$ and \nSpX$\times$; on a 25-frame benchmark sequence,
which \method covers with two overlapping windows, they are \nCostRatioSeq$\times$ and \nSpXSeq$\times$. Diffusion-VAS
spends \nSpDepth\,s on depth estimation alone, \sDepthX$\times$ the complete inference of \method, and \nSpDiff\,s on
denoising (means over the same run). The total parameter counts are of the same order (\nParamsOurs vs.\
\nParamsDvas), but \method trains only \nTrainOurs of its parameters (\sTrainPct), in the predictor (\sPredM), the LoRA adapters (\sLoraM) and the mask decoder (\sHeadM). The predictor is small next to the encoder, and the distilled world
models bring a pass down to \sTfB (ViT-B) and \nWmLTF TFLOPs (ViT-L) per window, where \method still recovers most of
its accuracy (Fig.~\ref{fig:analysis}a of the main paper).

\begin{table}[t]
\centering
\small
\setlength{\tabcolsep}{3pt}
\resizebox{\linewidth}{!}{%
\begin{tabular}{@{}lccccc@{}}
\toprule
Occluded & \method & Diff.-VAS & random-init & no pred. & frames \\
\midrule
0--10\% & 30.9 & 43.3 & 24.0 & 22.7 & 13,880 \\
10--20\% & 50.7 & 52.2 & 40.8 & 38.2 & 9,014 \\
20--30\% & 52.6 & 45.1 & 42.8 & 39.1 & 7,759 \\
30--40\% & 54.4 & 48.1 & 46.4 & 43.4 & 7,094 \\
40--50\% & 51.8 & 44.9 & 45.8 & 42.8 & 7,419 \\
50--60\% & 48.0 & 40.1 & 42.0 & 38.3 & 7,200 \\
60--100\% & 39.7 & 30.6 & 35.7 & 34.4 & 30,879 \\
\bottomrule
\end{tabular}}
\caption{\textbf{SAIL-VOS occluded IoU by occlusion ratio} over all occluded frames, with the values behind Fig.~\ref{fig:analysis}b of the main paper. The random-init variant trains the predictor from a random initialisation, and no pred.\ has no predictor.}
\label{tab:s_occ_sv}
\end{table}

\begin{table}[t]
\centering
\small
\setlength{\tabcolsep}{3pt}
\resizebox{\linewidth}{!}{%
\begin{tabular}{@{}lccccc@{}}
\toprule
Visibility & $<$0.25 & 0.25--0.5 & 0.5--0.75 & 0.75--1 & 1 \\
\midrule
\method & 41.5 & 77.7 & 91.9 & 99.0 & 99.7 \\
Diffusion-VAS & 24.5 & 43.4 & 89.0 & 97.1 & 97.9 \\
random-init predictor & 44.8 & 74.8 & 90.0 & 98.9 & 99.3 \\
no predictor & 45.8 & 76.5 & 89.0 & 98.7 & 98.8 \\
frames & 583 & 1,171 & 2,210 & 9,104 & 10,608 \\
\bottomrule
\end{tabular}}
\caption{\textbf{TAO-Amodal val AP$_{50}$ by visibility} (frames with the amodal box inside the image; values behind Fig.~\ref{fig:analysis}c of the main paper).}
\label{tab:s_occ_tao}
\end{table}

\paragraph{Occlusion level (\cref{tab:s_occ_sv,tab:s_occ_tao}).} On SAIL-VOS, \method leads in every bin from 20\%
occlusion upwards, by \nOcSvGapMin to \nOcSvGapMax points, and these bins hold \sOcSvShareHigh of the occluded frames.
The most occluded bin, 60--100\%, is also the largest (\sOcSvNTop frames); there, where amodal completion matters
most, \method reaches \nOcSvHighOurs against \nOcSvHighDvas. Below 20\% occlusion, Diffusion-VAS is ahead, most clearly below 10\%
(\nOcSvLowDvas vs.\ \nOcSvLowOurs). There the hidden part is often a thin sliver, finer than the token grid (a token
typically spans one eighth of the object's longer side), which pixel-level generation fits better. Both ablated models
stay below \method in every bin, so the pre-trained predictor helps across the whole range of occlusion. On
TAO-Amodal, all three variants of our model, which share the crop, the encoder and the mask decoder, are far ahead of
Diffusion-VAS when less than half of the object is visible, with at least \sOcTaoQMin and \sOcTaoHMin AP$_{50}$ in the two lowest bins against \nOcTaoQDvas and \nOcTaoHDvas. At such low visibility, the advantage on TAO-Amodal therefore comes
from these shared components rather than from the predictor.

\begin{figure}[t]
\centering
\includegraphics{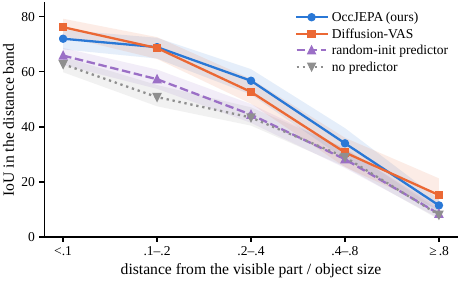}
\caption{\textbf{IoU by distance from the visible part} on all \nSvNset occluded SAIL-VOS val frames. IoU is computed within bands of distance from the visible mask, in units of object size (square root of the amodal area), and shading shows 95\% bootstrap CIs over videos.}
\label{fig:s_band}
\end{figure}

\paragraph{Distance from the visible part (\cref{fig:s_band}).} This is the mask-level counterpart of
Fig.~\ref{fig:interp} of the main paper. The predictor adds most between 0.1 and 0.4 object sizes from the visible
part (\nIaGainB and \nIaGainC IoU over no predictor in the bands 0.1--0.2 and 0.2--0.4). Diffusion-VAS is ahead in
the band nearest to the visible part and in the farthest one, and \method in the three bands between.

\begin{table}[t]
\centering
\small
\setlength{\tabcolsep}{3pt}
\resizebox{\linewidth}{!}{%
\begin{tabular}{@{}lcc@{}}
\toprule
Method & $|$IoU$(\hat a_t,\hat a_{t+1})-$IoU$(a_t,a_{t+1})|$ & IoU$(\hat a_t,\hat a_{t+1})$ \\
\midrule
\method & 7.15 {\scriptsize [5.98, 8.35]} & 74.5 {\scriptsize [69.9, 78.7]} \\
Diffusion-VAS & 3.96 {\scriptsize [3.21, 4.76]} & 77.0 {\scriptsize [70.9, 82.3]} \\
no predictor & 6.52 {\scriptsize [5.70, 7.37]} & 75.9 {\scriptsize [71.0, 80.1]} \\
ground truth & 0 & 80.0 \\
\bottomrule
\end{tabular}}
\caption{\textbf{Temporal stability on SAIL-VOS} (with $\hat a_t$ and $a_t$ the predicted and ground-truth amodal masks of frame $t$; all sequences, 24 consecutive pairs each, $\times100$; 95\% CIs over videos). Lower deviation means frame-to-frame changes closer to those of the ground truth.}
\label{tab:s_temporal}
\end{table}

\paragraph{Temporal behaviour (\cref{tab:s_temporal}).} The table measures how much the predicted amodal mask changes
between consecutive frames, relative to the change of the ground-truth mask. Diffusion-VAS generates all 25 frames
jointly, and its frame-to-frame changes are closer to those of the ground truth (deviation \nTsDevDvas vs.\ \nTsDevOurs). \method minus Diffusion-VAS is \sTsPairDev \sTsPairDevCI in deviation and \sTsPairIoU \sTsPairIoUCI in consecutive-frame IoU.
The predictor itself adds a little of this deviation (\sTsPairDevNone \sTsPairDevNoneCI against the model without it).
\method recomputes its crop in every frame and is trained for per-frame accuracy, on which it is ahead
(\cref{tab:s_sailvos,tab:s_tao}); temporally smoothed crops or a temporal loss are natural extensions.

\section{Qualitative results}
\label{sec:s_qual}

This section shows further SAIL-VOS, TAO-Amodal and Kubric sequences. In all figures, white is the visible part of the
predicted mask, magenta the completed hidden part, and the outline on the input frames is the given visible mask.

\begin{figure*}[t]\centering
\includegraphics{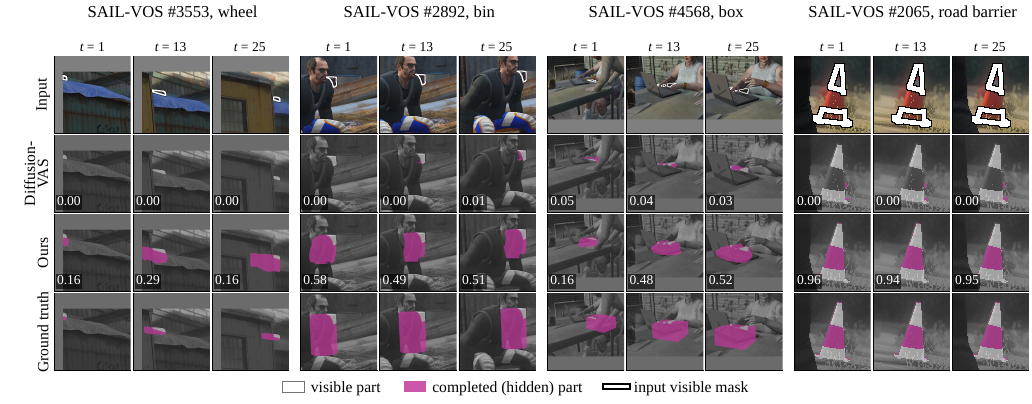}
\caption{\textbf{SAIL-VOS}, three frames per sequence. Numbers give the per-frame occluded IoU.}
\label{fig:s_sv}
\end{figure*}

\paragraph{More SAIL-VOS sequences (\cref{fig:s_sv}).} The first three sequences are heavily occluded (mean occlusion
\sQlRandOccMin to \sQlRandOccMax). In all three, Diffusion-VAS predicts little or no hidden part (sequence occluded IoU
at most \sQlRandDvasMax), while \method completes the object in every shown frame (sequence occluded IoU
\sQlRandOursMin to \sQlRandOursMax). The hidden objects are behind a tarp in the first sequence, behind the seated man
in the second (only a piece next to his head is visible), and under the man's hands in the third. The official metric does not count frames that a method misses in this way, and Diffusion-VAS misses \nDvasMissedPct of all occluded frames (\cref{tab:s_sailvos}).

\begin{figure}[t]\centering
\includegraphics{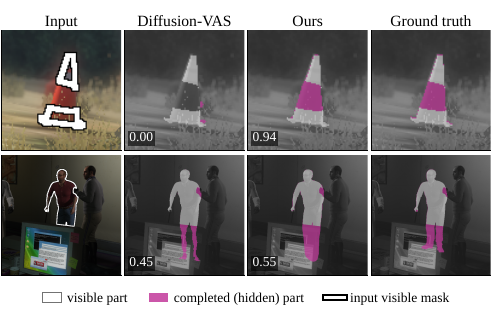}
\caption{\textbf{The first two SAIL-VOS sequences of Fig.~\ref{fig:features} of the main paper}, middle frame. Numbers give the occluded IoU of the frame.}
\label{fig:s_teaser}
\end{figure}

\paragraph{Feature-figure sequences (\cref{fig:s_teaser}).} In the first sequence, the middle of a traffic cone is
hidden and only its tip and base are visible. Diffusion-VAS predicts no hidden part, while \method fills the band
between the two visible pieces (sequence occluded IoU \sQlWinOurs vs.\ \sQlWinDvas). In the second sequence, both
methods complete the legs of a person standing behind a screen, and the completion of \method follows the ground truth
more closely.

\begin{figure*}[t]\centering
\includegraphics{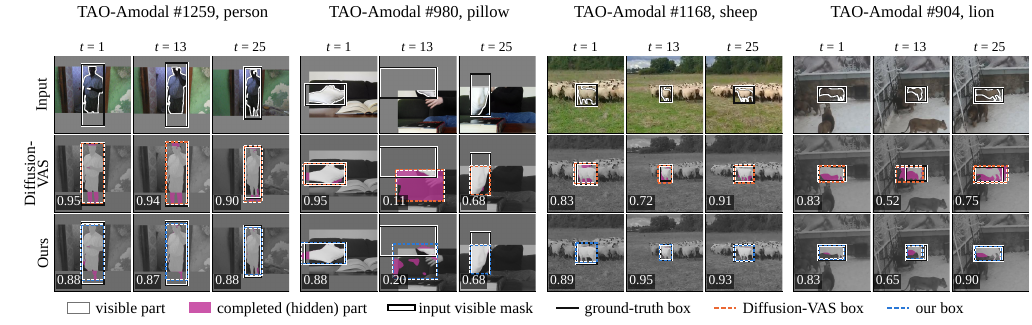}
\caption{\textbf{TAO-Amodal}, with the object leaving the frame (left two) and with in-frame occlusion (right two),
three frames per sequence. Numbers give the box IoU. Each crop follows the object, and grey marks the area outside the image.}
\label{fig:s_tao}
\end{figure*}

\paragraph{TAO-Amodal (\cref{fig:s_tao}).} The sequences come from real videos of indoor scenes (Charades) and of
animals (LaSOT). Under in-frame occlusion (right), the box of \method follows the hidden extent of the sheep in the
herd and of the lion more closely than that of Diffusion-VAS in every shown frame except the first lion frame, where
the two are tied. When the object leaves the image (left), both methods extend the box beyond the image border, and
Diffusion-VAS is ahead in four of the six shown frames; over all such val frames, \method is ahead by \sTaoOofValDiff
AP$_{50}$ (\cref{tab:s_tao}).

\begin{figure*}[t]\centering
\includegraphics{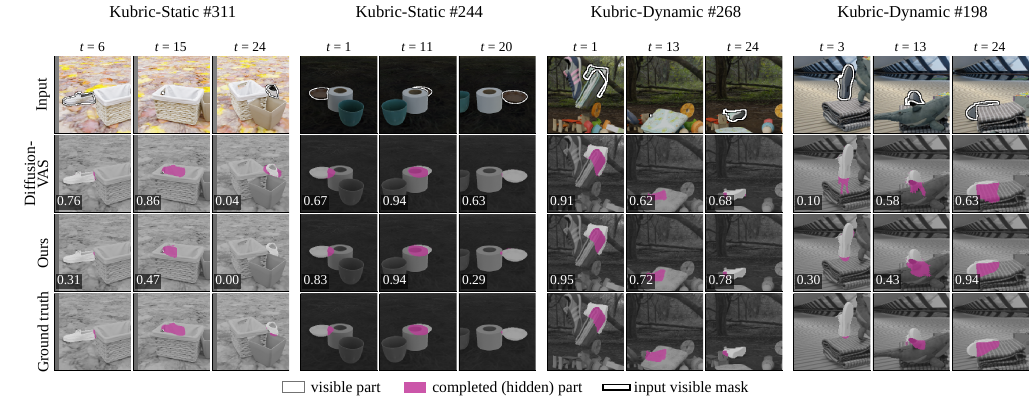}
\caption{\textbf{Kubric-Static (left two) and Kubric-Dynamic (right two)}, three frames per sequence. We run Diffusion-VAS on the same inputs. Numbers give the per-frame occluded IoU.}
\label{fig:s_kubric}
\end{figure*}

\paragraph{Kubric (\cref{fig:s_kubric}).} These sequences come from the test sets of \cref{tab:s_kubric}. On
Kubric-Dynamic, where the margin of \method is larger (\sKubDynOccDiff mIoU$_\text{occ}$), its completion is closer to
the ground truth in most of the shown frames, also when the hidden part changes shape as the objects move. On
Kubric-Static, where the average margin is \sKubStaOccDiff, the per-frame differences go both ways, as in these two
sequences.

\FloatBarrier
\par

\end{document}